\pdfoutput=1

\documentclass[11pt]{article}

\usepackage{acl}

\usepackage{times}
\usepackage{enumitem}
\usepackage{latexsym}
\usepackage{booktabs}
\usepackage{float}
\usepackage{amssymb}
\usepackage{placeins}
\usepackage[T1]{fontenc}

\usepackage[utf8]{inputenc}
\usepackage{tabularx}
\usepackage{array}     
\usepackage{microtype}
\usepackage{amsmath}
\usepackage{inconsolata}
\usepackage{dblfloatfix}
\usepackage{placeins}
\usepackage{graphicx}
\usepackage{natbib}
\usepackage[textwidth=2.1cm]{todonotes} 
\usepackage{relsize} 
\usepackage{mathtools}
\usepackage{booktabs}
\usepackage{multirow}
\usepackage{afterpage}
\usepackage{longtable}
\usepackage{float}
\usepackage{makecell}
\usepackage{xcolor}    
\usepackage{hyperref}  
\usepackage{etoolbox}
\usepackage{tikz}
\usepackage{caption}

\usetikzlibrary{positioning, arrows.meta, shapes.multipart, fit, backgrounds}

\newcolumntype{C}[1]{>{\centering\arraybackslash}p{#1}}
\newcommand{\maskedtok}{\text{\relsize{-1}{\sc mask}}}

\newcommand{\genmethod}{\textsc{mere}}
\newcommand{\gentest}{\textsc{merGe}}

\newif\ifhidecontent
\hidecontentfalse

\newif\ifhidecontenttwo
\hidecontenttwofalse 

\newcommand{\mlm}{M}

\title{\gentest: Minimal Expression-Replacement GEneralization\\Test for Natural Language Inference}

\author{
  Mădălina Zgreabăn \quad
  Tejaswini Deoskar \quad
  Lasha Abzianidze \\[2mm]
  Institute for Language Sciences, Utrecht University, The Netherlands \\[1mm]
  \texttt{\{b.m.zgreaban, t.deoskar, l.abzianidze\}@uu.nl}
}

\begin{document}
\maketitle
 \begin{abstract}
As many benchmarks have become saturated, it is increasingly important to create new datasets that evaluate the generalization capacity of current state-of-the-art models in reasoning.
However, creating high-quality reasoning datasets is challenging: manual construction is costly, and automatic generation is error-prone, with the community therefore relying on synthetic datasets with limited scope.  
In this paper, we propose the Minimal Expression Replacement GEneralization (\gentest) test, to evaluate the robustness and generalization of reasoning models. We use minimal and non-adversarial variants of existing evaluation datasets. First, we obtain high-quality variants automatically from the original instances using Masked Language Models (MLMs) for generation together with safeguarding filters, which we call the Minimal Expression Replacement (\genmethod) method.
We then apply the \gentest\ test to Natural Language Inference (NLI), a popular reasoning task, by using \genmethod\ on two popular existing NLI datasets. We evaluate multiple strong NLI models and LLMs, and the results indicate they generalize poorly:  
both struggle to consistently and correctly classify variants minimally different in form, but similar in reasoning, from the original ones. 
We also analyze how aspects of variant generation, such as word class and source MLMs, affect model performance.
\end{abstract}

\section{Introduction} 

\paragraph{} 

The challenge in the Natural Language Inference (NLI) task is to predict the inference relation between a premise $p$ and hypothesis $h$.
Current models achieve very high performance on NLI, but this performance has been shown to often depend on partial exploitation of heuristics learned from the training set \cite{gardner-etal-2020-evaluating,hupkes2023taxonomy, dutt-etal-2024-investigating}.
Thus, when tested on \textit{out-of-distribution} (OOD) datasets \citep{hupkes2023taxonomy, budnikov2025generalization, dutt-etal-2024-investigating, yang-etal-2023-distribution}, where test items differ from training ones in aspects that are not crucial to solving the task itself, such as text genres, models generalize poorly \cite{nie2019adversarial, verma-etal-2023-evaluating}, despite good performance on traditional in-distribution testsets of datasets like SNLI \citep{bowman2015large} or MNLI \citep{williams-etal-2018-broad}.

\begin{figure}[t!]
    \centering
    \includegraphics[clip, trim=3mm 96mm 310mm 5mm, width=\linewidth]{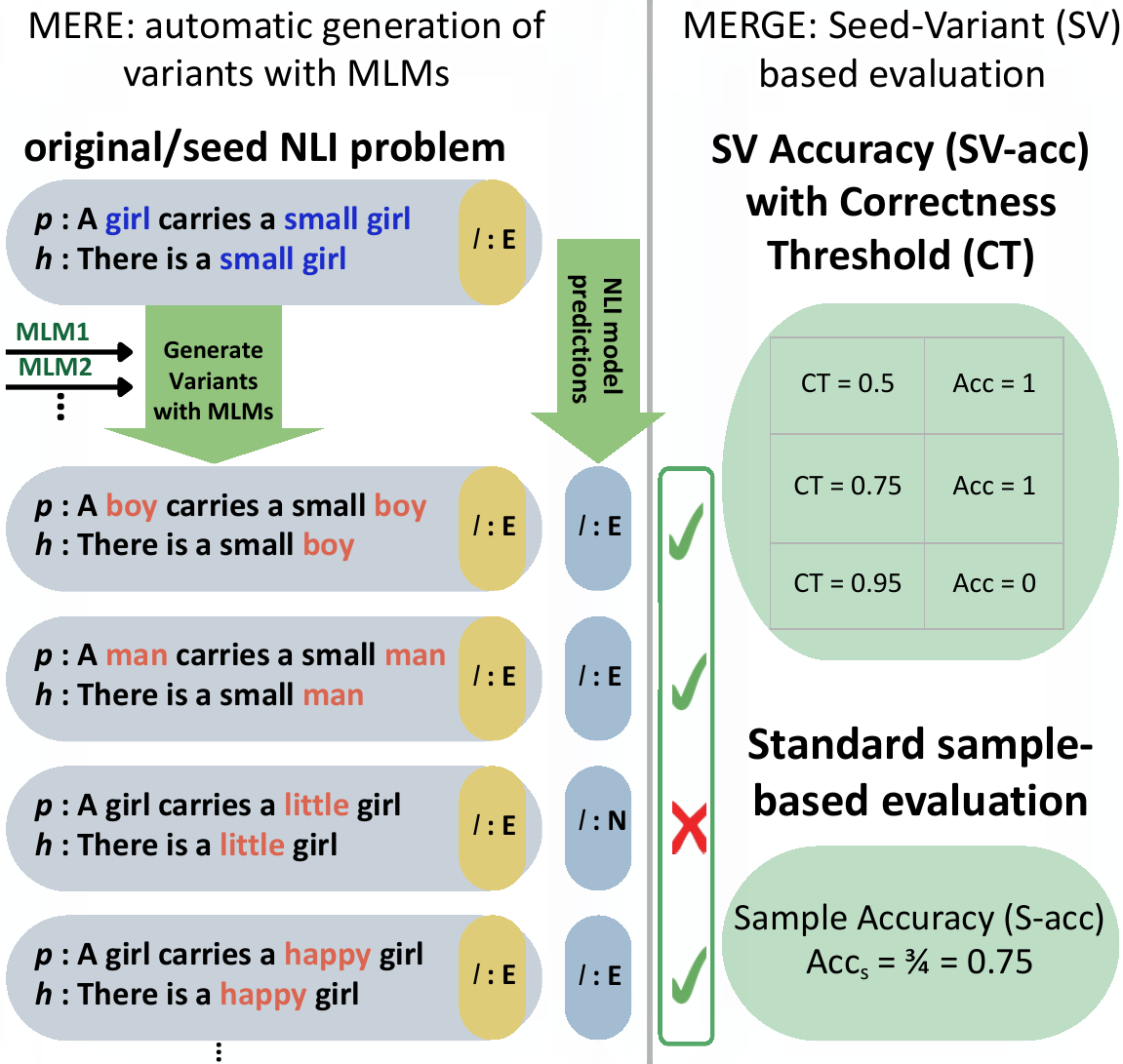}
    \caption{First, the \genmethod\ generation creates multiple label-preserving variants from a seed problem.
    Then \gentest\ evaluates a model on the seed based on its performance on the variants.
    Unlike the standard sample-based accuracy metric, \gentest\ uses a Seed-Variant (SV) accuracy with a correctness threshold, counting the seed as correctly classified if its variants are likewise classified correctly to at least the threshold degree.
    }
    \label{fig:metric}
\end{figure}

OOD NLI datasets are usually challenging when they don't retain common heuristics, such as word overlap \cite{rajaee-etal-2022-looking} or hypothesis-only artifacts \cite{gururangan-etal-2018-annotation,poliak-etal-2018-hypothesis,tsuchiya-2018-performance}. Additionally, performance also drops on adversarial datasets, designed to go against the training data distribution by resembling certain training samples, but with different labels \cite[][among others]{glockner-etal-2018-breaking, naik-etal-2018-stress, gardner-etal-2020-evaluating}.     
This raises the question of whether the decreased performance of models is due to a failure to generalize, or because OOD datasets are too disruptive, as argued by \citet{bowman-dahl-2021-will}. We therefore ask the question: can we create a minimal, non-adversarial generalization-challenge dataset that is ``friendly'' and deliberately retains such heuristics? Such a challenge set would test the ability of NLI models and LLMs to generalize on the task itself, without conflating it with additional challenges. 

In this paper, we propose a method of automatically generating such evaluation datasets from existing ones by creating \textit{minimal variants} of original `seed' problems (Minimal Expression Replacement, \genmethod).
\autoref{fig:metric} provides an overview of our methodology: the left side shows how \genmethod\ operationalizes minimality for an NLI problem. To form variants of the seed problem $\langle p,h \rangle$, MLMs are used to suggest contextually probable replacements for the open-class words shared between $p$ and $h$. The variants are designed to maintain the underlying reasoning of the seeds, while also preserving common heuristics, such as sentence length, that models can still exploit.

The right side in \autoref{fig:metric} shows the full methodology for evaluation of generalization on the generated variants (Minimal Expression-Replacement GEneralization, \gentest{}). In \gentest{} we use a pattern-based accuracy metric \citep{abzianidze-etal-2023-spacenli}, hereafter called Seed-Variant (SV) accuracy, instead of a usual accuracy measure (i.e., sample accuracy). Thus,  a model is evaluated not only on its predictions on seed problems, but also on its consistency, measured by predictions on corresponding variants of the seed problems.

We focus on the following questions: 
(i) how reliable is the methodology for automatically obtaining variants (\genmethod)?
(ii) how well do NLI models and LLMs generalize to minimal variants (\gentest)?  
(iii) how do factors of variant generation, like the word class of replacements or the source MLM, influence model performance? Our contributions and findings are as follows:
\vspace{-1.5mm}
\begin{enumerate}[itemsep=-3pt,leftmargin=*]
    \item A fully automated methodology for creating high-quality model-friendly test sets for generalization, applied to NLI.
    \item Our results reveal poor generalization in NLI, in both fine-tuned NLI models and LLMs, even under minimally changed datasets.
    \item We conduct extensive analyses that show verb replacements are difficult for models to generalize on, followed by nouns and adjectives. Surprisingly, the source MLM that generates the variant does not affect NLI models' performance, even those that share architecture.
\end{enumerate}

\section{Related Work}\label{relatedwork}

\begin{table*}[t]
\scriptsize
\center
\resizebox{\linewidth}{!}{
\begin{tabular}
{p{3.1cm}p{1cm}p{0.8cm}p{0.6cm}p{1.5cm}p{1cm}p{0.2cm}p{0.2cm}p{0.2cm}p{0.4cm}p{1cm}}
\toprule
\textbf{Study} &
\multicolumn{5}{c}{\textbf{Modification}} &
\multicolumn{4}{c}{\textbf{Preservation}} &
\textbf{Evaluation} \\

\cmidrule(lr){2-6}
\cmidrule(lr){7-10}

& \textbf{Type} & \textbf{Procedure} & \textbf{Unit} & \textbf{Dataset} & \textbf{Validation}
& \textbf{M} & \textbf{R} & \textbf{S} & \textbf{WO}
& \\
\midrule

\citet{li-etal-2020-linguistically}
& Multiple & Auto. & $p$ & SNLI; MNLI & H$_p$
& $\pm$ & $\pm$ & $-$ & $-$
& G$_\textit{Ms}$; O$_\textit{Ms}$ \\
\midrule

\citet{glockner-etal-2018-breaking}
& Replace & Auto. & $h$ & SNLI & H$_f$
& $-$ & $\pm$ & $+$ & $-$
& G$_m$ \\
\midrule

\citet{verma-etal-2023-evaluating}
& Paraphrase & Auto. & $p$; $h$; $p\mathord{\wedge}h$ & Pascal RTE1-3 & H$_f$
& $+$ & $+$ & $-$ & $-$
& O$_\textit{ms}$ \\
\midrule

\citet{srikanth-etal-2024-often}
& Paraphrase & Mix & $h$ & $\alpha$-NLI; $\delta$-NLI & H$_f$
& $+$ & $+$ & $-$ & $-$
& Ms$_\textit{Ms}$; G$_\textit{Ms}$ \\
\midrule

\citet{arakelyan-etal-2024-semantic}
& Paraphrase & Auto. & $h$ & SNLI; MNLI; ANLI & H$_p$
& $+$ & $+$ & $-$ & $-$
& O$_m$ \\
\midrule

\citet{petrov2025superficial}*
& Multiple & Auto. & $h$ & SNLI & N/A
& $-$ & $-$ & $-$ & $-$
& G$_m$ \\
\midrule

\citet{kaushik2020learningdifferencemakesdifference}
& Multiple & Man. & $p$; $h$ & SNLI & H$_f$
& $-$ & $-$ & $-$ & $-$
& G$_m$ \\
\midrule

\citet{srikanth2025nlimicroscopeatomichypothesis}
& Decompose & Auto. & $h$ & SNLI; $\delta$-NLI & Mix
& $+$ & $+$ & $-$ & $-$
& G$_m$; O$_\textit{Ms}$ \\
\midrule

\textbf{\gentest\ } (our method)
& \textbf{Replace} & \textbf{Auto.} & \textbf{$p\mathord{\wedge}h$} & \textbf{SNLI; MNLI} & \textbf{LM}
& $\mathbf{-}$ & $\mathbf{+}$ & $\mathbf{+}$ & $\mathbf{+}$
& \textbf{G$_m$; G$_\textit{Ms}$} \\
\bottomrule

\end{tabular}}
\caption{OOD NLI datasets classified by the type of modifications used to obtain variants (Type); the manual or automatic procedure (Procedure); the unit modified (Unit) and its origin dataset (Datasets -- Pascal RTE1-3 \cite{dagan2005pascal}; $\alpha$-NLI \cite{bhagavatula2019abductive}; $\delta$-NLI \cite{rudinger-etal-2020-thinking}); and the manual or automatic method of validation (Validation). Other columns show 
the preservation of the meaning of the original problem (M), its reasoning (R), syntax (S) or word overlap (WO), and how variants were used in evaluation (Evaluation). \S \ref{relatedwork} describes what each element of the columns means. * indicates that the variants were used for fine-tuning models. All studies use labels: entailment, contradiction, and neutral.}
\label{classbench}
\end{table*}

We compare various OOD datasets that have been used in prior NLI work to facilitate comparison with our own work, and classify them in \autoref{classbench}, according to various criteria as below: 

\paragraph{Modification Type} Modifications in the OOD dataset were made by \textit{decomposing} or \textit{paraphrasing} original problems, by \textit{replacing} their words, or by combining \textit{multiple} operations (Column Type). 
\vspace{-0.3em}
\paragraph{Modification Procedure} Modifications  were made \textit{automatically}, \textit{manually}, or through a \textit{mix} of both (Column Procedure).

\paragraph{Modified Unit} Modifications were applied only to $p$, only to $h$, or across $p$ and $h$ ($p\mathord{\wedge}h$) (Column Unit).

\paragraph{Validation} Modified problems were validated manually by having a partial or full set of variants annotated (H$_p$, and H$_f$), through a mix of automatic and manual methods (\textit{Mix}), or not at all (\textit{N/A})(Column Validation).
\paragraph{Meaning, Reasoning, Word Overlap and Syntax} The meaning of the problem (M), underlying reasoning (R), syntax (S), or word overlap of $p$ and $h$ (WO)\footnote{WO preservation requires modifying $p\mathord{\wedge}h$ at the same time.} were \textit{preserved} ($+$), \textit{changed} ($-$), or a \textit{mix} of both ($\pm$) (Columns under Preservation).

\paragraph{Evaluation} Predictions on modified problems $m$ were evaluated either \textit{individually}, against the gold label (\textit{G$_m$}) or the prediction on the initial problem (\textit{O$_m$}), or \textit{jointly}, against the gold label (\textit{G$_\text{Ms}$}), the prediction on the original problem (\textit{O$_\text{Ms}$}) or against each other (\textit{Ms$_\text{Ms}$}) (Column Evaluation).\\

The performance drops in previous OOD NLI datasets cannot be attributed fully to poor generalization, as they were constructed with non-syntax-preserving constructions that might be more challenging for models \citep{li-etal-2020-linguistically}, or changed the lexical overlap of $p$ and $h$ due to paraphrased or decomposed NLI problems. Preserving word overlap between $p$ and $h$ has been partially explored previously by \citet{glockner-etal-2018-breaking}, but their replacements did not preserve reasoning and favored variants' plausibility at the expense of lexical diversity. Other shortcomings of previous variants' datasets concern their automatic generation, previously criticized for potentially (i) biasing datasets in favor of models deployed \cite{li-etal-2020-linguistically, gardner-etal-2020-evaluating}; and (ii) constructing implausible datasets \citep{dutt-etal-2024-investigating}. However, manual variant dataset creation is very time-consuming and unlikely to keep pace with the rate at which new models are developed,  making a reliable automatic generation methodology necessary. 

We propose a method that automatically creates high-quality plausible variants by replacing shared words of $p$ and $h$ with felicitous alternatives. 
Our method is the only one designed to preserve underlying reasoning, lexical overlap of $p$ and $h$, and their syntax. 
Our method also limits the implausibility of constructed variants by selecting replacements that language models deem {\it at least as probable} in context as the words they replace (described in detail in \autoref{metho}). Additionally,  \genmethod\ can also vary the level of lexical diversity by including more or fewer MLMs. 

\section{Methodology}\label{metho}

\genmethod\ takes an original seed NLI problem $\langle p,h,l\rangle$, labeled with $l$, and generates a set of replacement words $r$ from a set of MLMs for each open-class word $o$ shared between $p$ and $h$. Then, it quality-filters the replacements, unifies them across MLMs, and restricts them to those compatible with both $p$ and $h$. Finally, \genmethod\ generates label-preserving variants $\langle p_{i},h_{i},l\rangle$ by replacing $o$ with $r$.

\begin{figure}[t!]
    \centering
    \includegraphics[clip, trim=00mm 70mm 340mm 00mm, width=\linewidth]{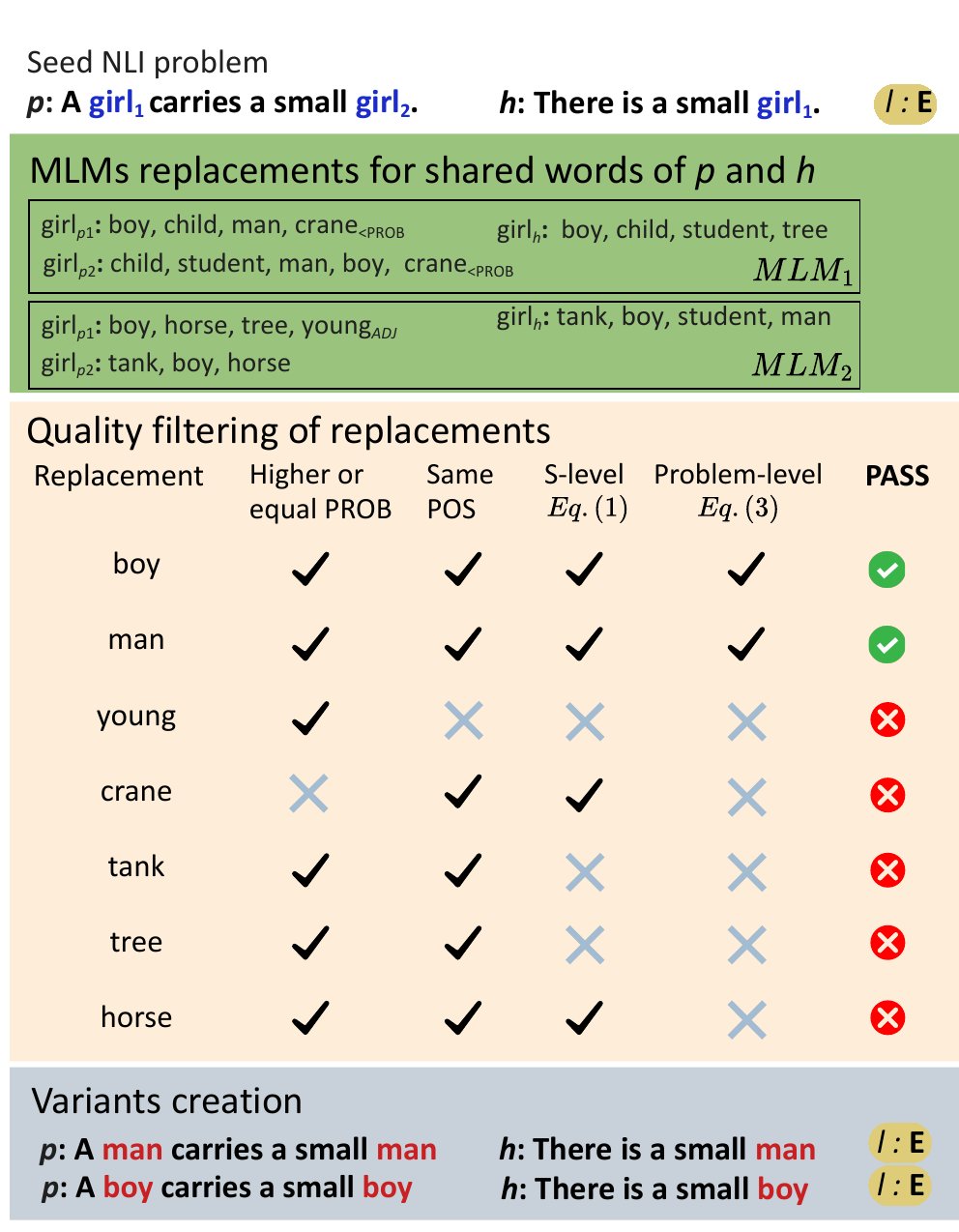}
    \caption{Generation of NLI variants with MLM replacements. Replacements must be at least as probable as $o$ (Higher or equal PROB), preserve the class of $o$ (Same POS), and be suggested by the same MLM for all multiple occurrences at the sentence level. At the problem level, replacements must be suggested for both $p$ and $h$, although they might come from different MLMs.}

    \label{fig:examplefiltering}
\end{figure}

First, the set $r$ is collected from the set of MLMs $\mathcal{M} = \{\mlm_1,\ldots,\mlm_n\}$ for a given sentence $S=(s_1,\ldots,s_k)$ (either $p$ or $h$), an MLM $\mlm_j$, and an open-class word $o=s_i$, indicated by $R_{j}(S, i)$, as shown in the upper part of \autoref{fig:examplefiltering}.


Second, $r$ is filtered with safeguarding constraints (i)-(iii) to create $R_{j}(S, i_{>}^c)$: \\
(i) $r$ has a higher probability than $o$ in the masked context $S=(s_1,\ldots,s_i/\maskedtok,\ldots,s_k)$ under $\mlm_j$, marked with $i_>$ ;\\ 
(ii) $r$ and $o$ are of the same word class\footnote{The possible word classes are nouns, verbs, adjectives, or adverbs; note that if $s_i$ is not part of the $\mlm_j$ vocabulary, the set of replacements will be empty.} in $S$ at position $i$, marked with $i\overset{c}{}$ ;\\
(iii) $r$ does not occur in $S$\\ 
Constraints (i) and (ii) ensure that replacements $r$ are \emph{more} felicitous in $S$ than $o$, according to MLM probability, while preserving the syntax of $S$. For example, ungrammatical or implausible variants that could have resulted from replacing the shared noun `girl' in \autoref{fig:examplefiltering} with a lower-probability replacement in the context, such as `crane', or with a replacement with a different open class, such as `beautiful', are avoided by (i) and (ii). Constraint (iii) also avoids label non-preserving changes. For example, substituting a word like `boy' with `poodle' in a problem with $p$ `two poodles and a boy swim' and $h$ `only one boy swims' would change the initial entailment to contradiction\footnote{
While (iii) is a strong filter, variants with incorrect inference labels remain possible, although improbable, e.g., the replacement of `boy' with `dog' in the $p$ `two poodles and a boy swim'; $h$ `only one boy swims'.}. 

After filtering, if a word $o$ occurs multiple times in $S$, we define the replacement set from $\mlm_j$ as: 
{
\setlength{\abovedisplayskip}{2pt}
\setlength{\belowdisplayskip}{2pt}
\setlength{\abovedisplayshortskip}{2pt}
\setlength{\belowdisplayshortskip}{2pt}
\begin{equation}
\label{eq:within_sentence}
R_{j}(S, o_{>}^c)
=
\bigcap_{o=s_i} R_{j}(S, i_{>}^c)
\end{equation}
}
\noindent For example, \autoref{fig:examplefiltering} shows the exclusion of the replacement word `tank' as it is not suggested for both sentence-level occurrences of `girl' in $p$.

Afterward, for a set of MLMs $\mathcal{M}$, we define a set of replacements for a word $o$ in $S$ as:
{
\setlength{\abovedisplayskip}{2pt}
\setlength{\belowdisplayskip}{2pt}
\setlength{\abovedisplayshortskip}{2pt}
\setlength{\belowdisplayshortskip}{2pt}
\begin{equation}
\label{eq:p_h_unite}
R_{\mathcal{M}}(S, o_{>}^c)
=
\bigcup_{\mlm_j \in \mathcal{M}}
R_{j}(S, o_{>}^c)
\end{equation}
}
\noindent where replacements are validated by the same MLM at each occurrence of $o$ in $S$. 

Finally, for a sentence pair $\langle p,h\rangle$ and $o \in p \cap h$, we define a set of replacements from $\mathcal{M}$ as:
{
\setlength{\abovedisplayskip}{2pt}
\setlength{\belowdisplayskip}{2pt}
\setlength{\abovedisplayshortskip}{2pt}
\setlength{\belowdisplayshortskip}{2pt}
\begin{equation} \label{eq:p_h_intersection}
R_{\mathcal{M}}(\langle p,h \rangle, o_{>}^c) = \bigcap_{S \in \{p,h\}} R_{\mathcal{M}}(S, o_{>}^c)
\end{equation}
}

\noindent Note that intersection occurs when $r$ needs to be compatible with two simultaneous positions of $o$ (from multiple occurrences or sentences). Union, by contrast, expands the pool of $r$ to maximize the potential final set of intersected $r$.\footnote{We tested whether relaxing several criteria could lead to a `friendlier' methodology, such as removing safeguards (i), (ii), applying no filters, or using union for $p$ and $h$ in \autoref{eq:p_h_intersection}. However, our current methodology yields the best model results, as shown in \autoref{filtering_criteria}, making it the `friendliest'.}

Finally, the variants $\langle p_{ij},h_{ij},l\rangle$ of $\langle p,h,l\rangle$ are obtained by replacing the original shared words $o_i \in p\cap h$ with corresponding replacements $r_{ij} \in R_{\mathcal{M}}(\langle p,h \rangle, o_{i>}^c)$:

\smallskip
\centerline{
$p_{ij} = p[o_i/r_{ij}], h_{ij} = h[o_i/r_{ij}]
$}
\smallskip
\noindent as also shown in the last part of variant creation in \autoref{fig:examplefiltering}.
We will use $R^{d = m}$ to denote a subset of size $m$ of the replacement set $R$.
We call $d$ a degree of inflation.
If there are $k$ words shared between $p$ and $h$, and for each word we have a set of replacements with the inflation degree of $d$, then the total number of generated variants will be $k \times d$.

\section{Experimental Setup}\label{experiments}

We will now describe the various experimental choices made in building our variant dataset.

\paragraph{Replacement generation} 
Using SNLI test and the matched and mismatched MNLI-m/-mm development sets, for each problem sharing at least a noun, verb, or adjective\footnote{Adverbs were not included due to their low number of seed problems. SNLI and MNLI were chosen to maximize the pool of possible seed problems, being the largest NLI datasets.} across $p$ and $h$, we generated 200 replacements using each of the following MLMs: BERT \citep{DBLP:journals/corr/abs-1810-04805}, RoBERTa \citep{liu2019robertarobustlyoptimizedbert}, ALBERT \citep{DBLP:journals/corr/abs-1909-11942}, ELECTRA \citep{clark2020electrapretrainingtextencoders}, BART \citep{lewis2019bartdenoisingsequencetosequencepretraining}. We used the base and large sizes for each MLM except ALBERT (details in Appendix \autoref{tab:_MLM_overview}).

\paragraph{Automatic quality filtering} Replacements were kept if they were: i) of higher probability than $o$ in the masked context%
\footnote{
note that $r$ rarer than $o$ are not excluded by this constraint: we found that $o$ are more frequent than corresponding $r$ in large corpora, with an $\approx$0.5 higher Zipf frequency in \texttt{wordfreq} \citep{robyn_speer_2022_7199437}.}, ii) the same word class as $o$ , 
iii) different from words in $p$ and $h$ , 
iv) supported by at least one MLM.
Most replacements were filtered out by (i), with only 10--30\% remaining; see Appendix \autoref{avrgredc} for details.
After quality filtering (see below), we retained only seed problems that had at least 20 variants.%
\footnote{The datasets obtained are publicly available at \href{https://github.com/mdzgr/MERGE}{https://github.com/mdzgr/MERGE}.}   

\paragraph{Manual Quality Validation} The validation process had two stages. First, we manually validated 100 randomly sampled variants per open word class and dataset, for a total of 900 problems. We annotated the variants considering two aspects: the \textbf{reasoning} preservation of the original inference relation (R), and the \textbf{fluency} of the replacement (F).
F also penalizes ungrammatical replacements, regardless of the grammaticality of a seed problem. 
We scored F and R from 1 to 5,%
\footnote{For example, `Colorless green ideas sleep furiously' entailing "Green thoughts is angrily sleeping" should be assessed with 1 for F and 5 for R, in short F1-R5.
}
and kept only variants with F$+$R$\geq9$.

Two authors independently annotated 300 SNLI problems first, with a high inter-annotator agreement (Cohen’s $\kappa = .80$, $.77$, and $.89$ for noun, adjective, and verb variants, respectively). All BART replacements were excluded given they contributed more to forming bad variants, as shown in the distribution of good and bad SNLI variants in Appendix \autoref{fig:combined-threshold-9}. We then repeated this first validation phase for MNLI-m/-mm, where one author annotated all 600 examples. These variants, which already excluded BART replacements, had all good scores ($\geq 9$) except for one with a lower fluency score.

In the second annotation stage, one author annotated 100 variants per dataset, across classes. All sampled variants for MNLI-m, 98\% for MNLI-mm, and 91\% for SNLI had a good F+R score. No variant received a rating with an R score below 5 or was judged ungrammatical\footnote {The difference in grammaticality between seed problems and their variants is only $\approx$0.03 for all datasets, as revealed by a RoBERTa CoLA-trained model (\texttt{textattack/roberta-base-CoLA}).}. \genmethod\ thus has a substantially lower error rate than previous work, allowing us to retain more of the automatically created dataset. For comparison, \citet{verma-etal-2023-evaluating} reported an $\approx$44\% error rate. 
Consult Appendix \autoref{annotation_gudidelines} for the full validation guidelines, statistics on diversity of replacements, and an additional annotated variant sample with an LLM.  

\begin{table}[!t]
\centering
\footnotesize
\setlength{\tabcolsep}{2.5pt}

\begin{tabular}{l l r C{0.45cm} C{0.45cm} C{0.45cm} c}
\toprule
\textbf{Dataset} &
\textbf{Class} &
\textbf{\# Seeds} &
\multicolumn{3}{c}{\textbf{Gold label (\%)}} &
\textbf{Unique} \\
\cmidrule(lr){4-6}
& & & \textbf{N} & \textbf{C} & \textbf{E} & \\
\midrule

\multirow[t]{4}{*}{SNLI}
 & Var$_\textit{N}$ & 1808 & 29 & 20 & 50 & 67 \\
 & Var$_\textit{V}$ & 506  & 30 & 18 & 50 & 66 \\
 & Var$_\textit{A}$ & 259  & 32 & 22 & 44 & 63 \\
 & Var & \textbf{2222} & \textbf{30}
 & \textbf{20} & \textbf{49} & \textbf{77} \\

\midrule

\multirow[t]{4}{*}{MNLI-m}
 & Var$_\textit{N}$ & 3058 & 25 & 31 & 43 & 86 \\
 & Var$_\textit{V}$ & 648  & 22 & 27 & 50 & 69 \\
 & Var$_\textit{A}$ & 703  & 26 & 23 & 49 & 72 \\
 & Var & \textbf{3663} & \textbf{25}
 & \textbf{30} & \textbf{44} & \textbf{95} \\

\midrule

\multirow[t]{4}{*}{MNLI-mm}
 & Var$_\textit{N}$ & 3452 & 25 & 31 & 43 & 89 \\
 & Var$_\textit{V}$ & 683  & 21 & 27 & 51 & 68 \\
 & Var$_\textit{A}$ & 779  & 23 & 28 & 48 & 73 \\
 & Var & \textbf{4016} & \textbf{24}
 & \textbf{30} & \textbf{45} & \textbf{103} \\

\bottomrule
\end{tabular}
\caption{Number of final seed problems per open word class (\#Seeds) and dataset, 
with gold-label percentages, and the average number of unique variants per problem across all subsamples of Var (Unique). \#Seeds is non-summative, as problems with replacements from multiple word classes are counted only once in Var.}
\label{section4.table2}
\end{table}
\paragraph{Final variants} 
For each dataset, we randomly sampled only 20 variants per open word class per seed problem, to avoid overpopulating the dataset with variants from `productive' seeds, i.e. seeds with a lot of generated variants. 
We repeated this subsampling 10 times to avoid randomization artifacts and formed the Var dataset, comprising these 10 subsamples for each open word class.  Table~\ref{section4.table2} shows the number of seeds for each Var, alongside their label distributions.

\paragraph{Evaluation Metrics} 
We evaluate models with two metrics: standard accuracy, where each variant or seed is an individual evaluation point (sample accuracy, S-acc), and SV-acc, borrowed from \citet{abzianidze-etal-2023-spacenli}. Under SV-acc, a seed problem 
is considered correctly classified only if a minimum proportion of its variants are as well; see \autoref{fig:metric}. We call this proportion the correctness threshold (CT).
If models generalize well, their SV-acc\footnote{Note that for Var the SV-acc for a CT is the averaged SV-acc for that CT across all its 10 subsamples.} over variants should not substantially fall behind their Seed S-acc. To quantify this, we define the matching correctness (MC) threshold as the CT at which a model's SV-acc over variants matches its S-acc over seed problems. 
\FloatBarrier

\begin{figure*}[!htbp]
\vspace{-5mm}
  \includegraphics[
    width=\textwidth
    ]{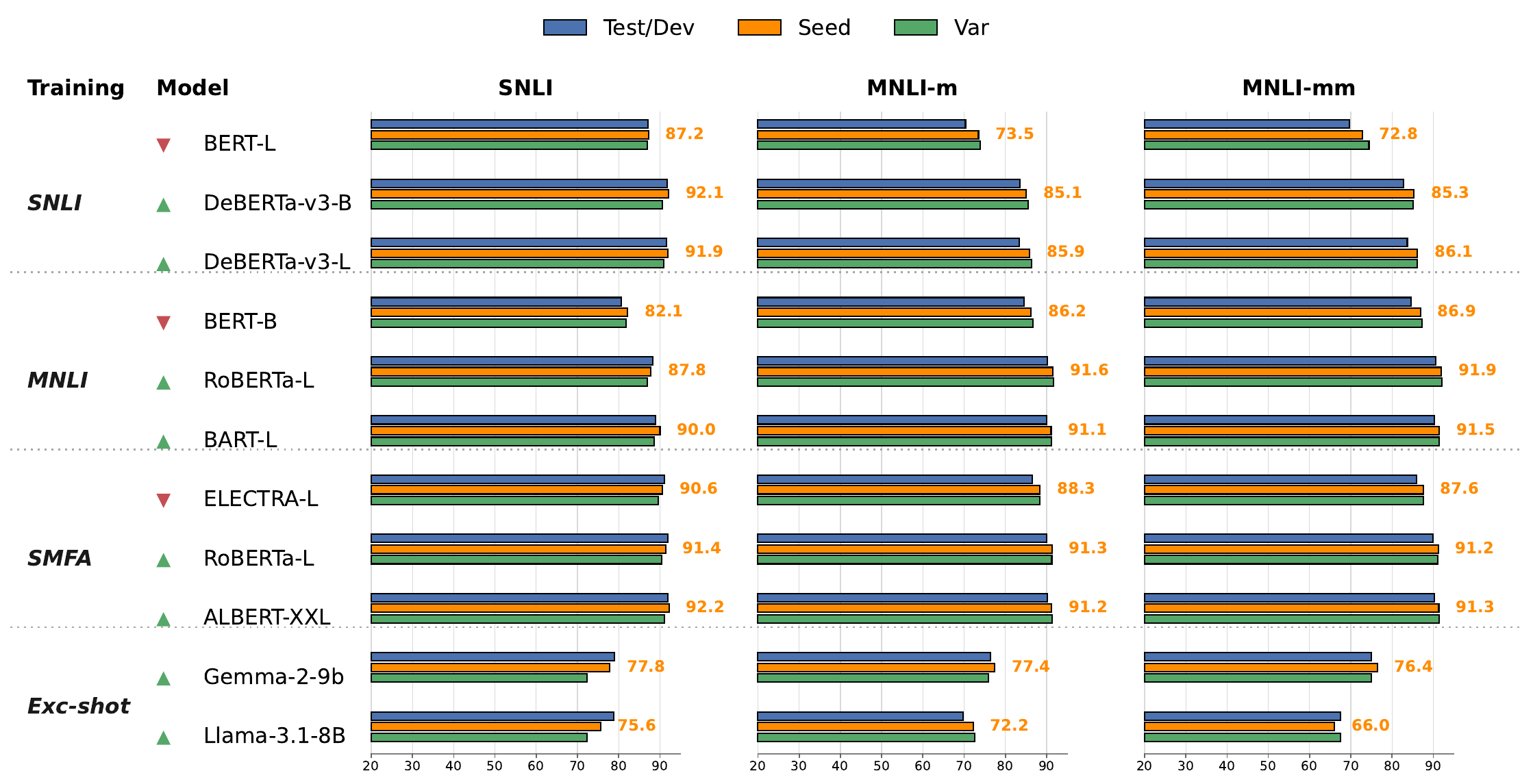}
    \vspace{-4mm}
    \caption{S-acc of the two best and the worst models trained either on SNLI, MNLI, SMFA, or, in the case of LLMs, prompted with examples exclusively chosen from the development set of SNLI and the train set of MNLI. Test/Dev is the S-acc on the original test problems for SNLI and dev problems for MNLI, while Seed and Var show the S-acc on seed problems, exactly shown by the orange numbers, and variant problems, respectively.}
    \label{fig:figure_3}
\end{figure*}

\paragraph{NLI Models} We evaluated several models fine-tuned on NLI to observe how NLI fine-tuning affects performance: BERT, RoBERTa, DeBERTa \citep{he2021debertadecodingenhancedbertdisentangled}, BART, ALBERT, ELECTRA, XLNet \citep{DBLP:journals/corr/abs-1906-08237}, OPT \citep{zhang2022opt}, and GPT-2 \citep{radford2019language}, fine-tuned on SNLI, MNLI, or a combination of SNLI, MNLI, FEVER \citep{nie2019combining}, and ANLI--SMFA. Full details of the models used are in Appendix \autoref{tab:appendix-model-cards}.  
\paragraph{LLMs} We also evaluated two LLMs: Gemma-2-9b \cite{gemma_2024}, and Llama-3.1-8B \cite{dubey2024llama}. 
For their evaluation, we randomly sampled 500 seed problems and their corresponding variants per dataset, picked to preserve the original seed distribution and Seed S-acc. We evaluated them in a multiple-choice task setting following the prompting strategy from \citet{madaan2024lostinferencerediscoveringrole} with six-shot%
\footnote{Six is the smallest even number above the 5 few-shot setting after which performance gains diminish in \citet{madaan2024lostinferencerediscoveringrole}.}
prompting examples either exclusively sampled from the dataset corresponding to the variants (SNLI-dev or MNLI-train), or as a 3+3 mix from both datasets. Lastly, we selected the model's label prediction based on its highest logit. Section \S \ref{prompting} of the Appendix shows additional details on seed subsampling, prompting, or few-shot examples.

\section{Results}\label{results}
\subsection{Do models generalize to variants?}\label{5.1}

We found that while models achieve high S-acc on individual variants, their performance drops significantly under SV-acc as we increase the Correctness Threshold (CT). 

\begin{figure}[!t]
    \centering
    \includegraphics[clip, trim=0mm 0mm 0mm    0mm,width=\columnwidth]{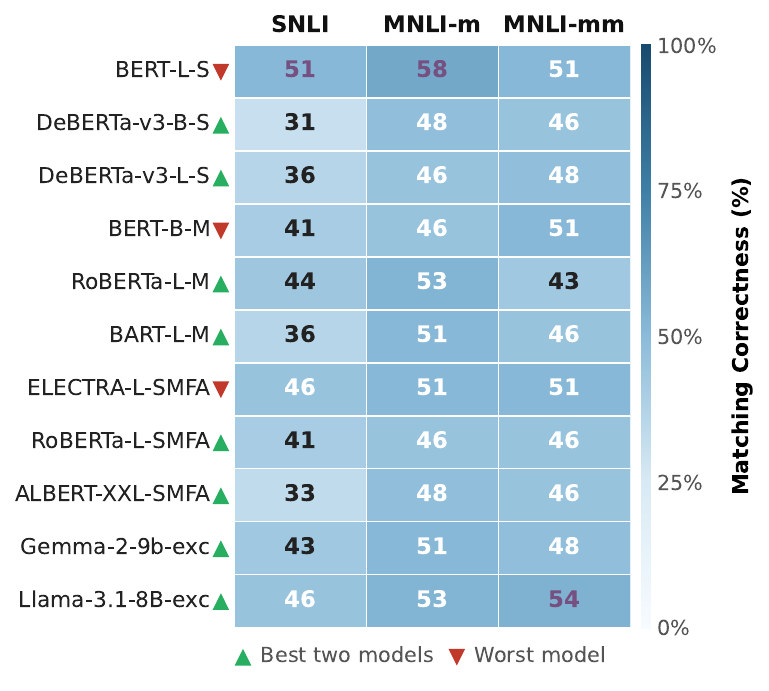}
    \caption{Matching Threshold, MC (the CT at which a model's SV-acc on variants matches its Seed S-acc), across models and datasets. 100\% MC would indicate perfect generalization, i.e., accuracy on all variants is the same as accuracy on seed. The last acronym in the model names specifies the datasets the models were fine-tuned on, or the type of prompting used. Best MCs are marked in purple, while MCs under 45 are in black.}
    \label{fig:heatmap}
\end{figure}

\autoref{fig:figure_3} shows results for the two best-performing and the worst-performing model per training data category\footnote{Selected by their averaged S-acc on test problems for SNLI and S-acc on dev problems for MNLI. Results for all models are in the Appendix, \autoref{fig:figure_15}.}. Their S-acc on individual variants is close to that on seed problems, as shown by the green and orange bars.
In contrast, the matching correctness (MC) heatmap (\autoref{fig:heatmap}) shows that, across datasets, most models cannot maintain their Seed S-acc when correctness is required on more than $\approx$50\% of variants for each seed. Above a CT of $\approx$60\%, performance drops across models and datasets, especially at higher thresholds.
Llama-3.1-8B-exc is the model that generalizes best on variants from MNLI-mm, i.e., with an MC of $\approx$54\%, while BERT-L-S has the highest MC for MNLI-m and SNLI, i.e., 58\% on MNLI-m.
\begin{figure*}[!b]
  \centering
  \includegraphics[width=\textwidth]{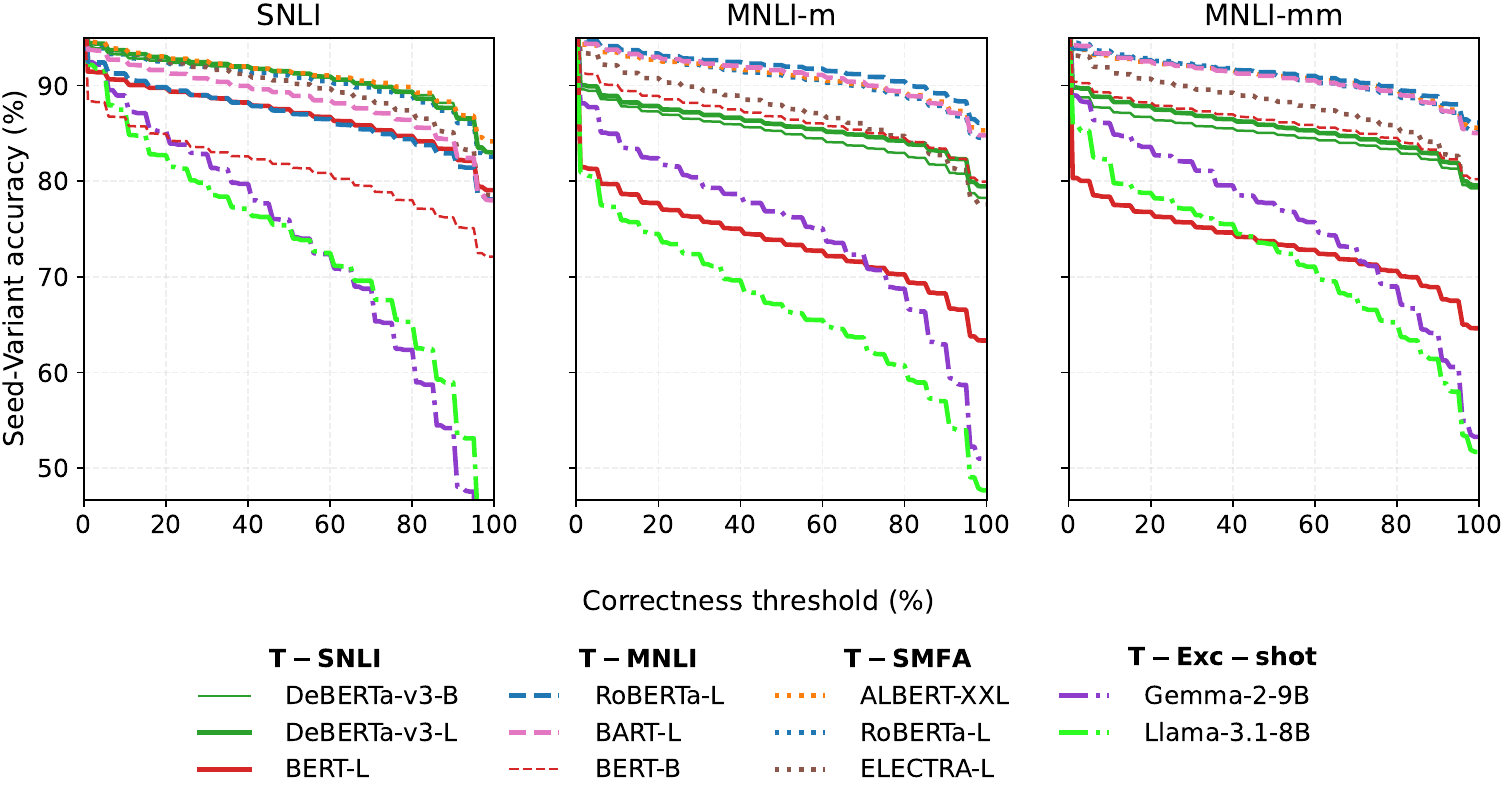}
  \caption{SV-acc vs. CT on variants from  SNLI and MNLI-m/-mm. The legend headers show the datasets the NLI models were fine-tuned on, i.e., SNLI, MNLI, or a combination of NLI datasets e.g., SMFA, or, for LLMs, the few-shot source: exclusively drawn from the source dataset of the variants (SNLI-dev for SNLI; MNLI-train for MNLI-m/-mm).}
  \label{fig:VA_curves_selected_models}
\end{figure*}
\autoref{fig:VA_curves_selected_models} further illustrates the models and their performance on different CTs. The differences between models' SV-acc over variants and Seed S-acc were shown to be statistically significant across models starting from the 60\% CT\footnote {Shown via a paired t-test (SNLI with $p=0.0027$, MNLI-m with $p=0.0003$, and MNLI-mm with $p=0.0001$).}. A particularly steep decline is seen in \autoref{fig:VA_curves_selected_models} for LLMs in the green and purple lines, even though Llama-3.1-8B-exc had the highest MC for MNLI-mm.\footnote{See description of selecting models for presentation, full model results and statistical analysis in the Appendix.} In fact, LLMs generally perform worse than all other NLI models, with S-acc lower on test or dev problems by as much as 20\%, as also shown previously by \citet{madaan2024lostinferencerediscoveringrole}. 

\begin{figure*}[!htbp]
  \small
  \centering
\includegraphics[width=\linewidth]{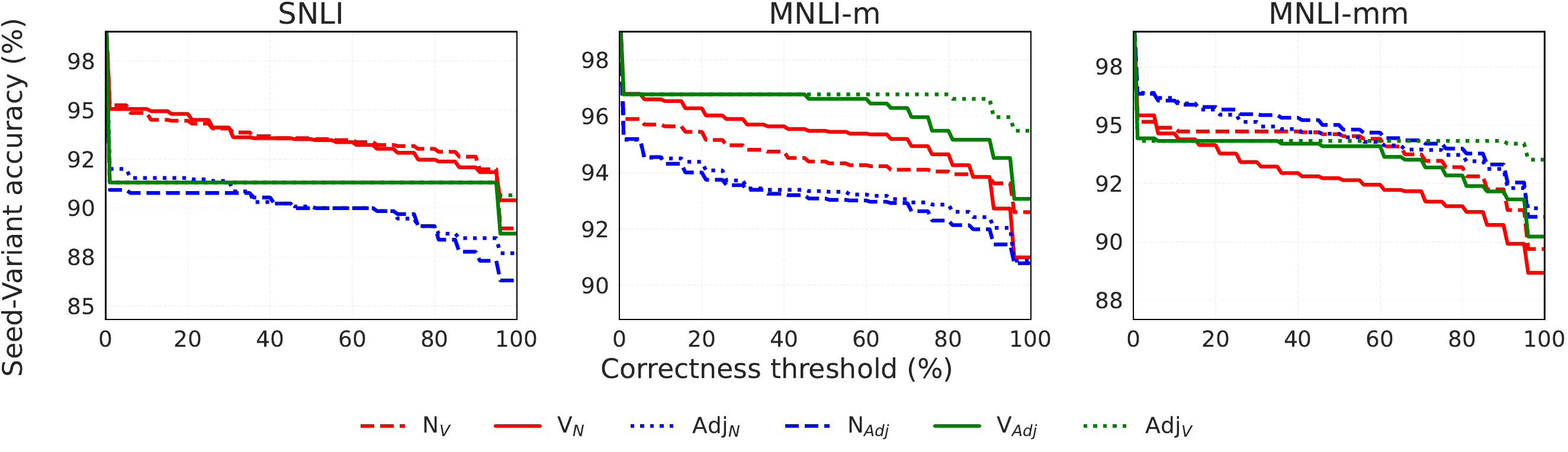} 
  \vspace{-5mm}
  \caption{SV curves of ALBERT-XXL on seed problems sharing at least two out of the three open classes.
  The baseline letters in the dataset names (N, V, Adj) show which open-class was replaced to form variants.}
  \label{nvnequal}
\end{figure*}

Models generalize better on MNLI-m/-mm variants than on those from SNLI, as shown by the higher MCs in \autoref{fig:heatmap}. This may be due to differences in seed difficulty: MNLI seeds are easier than their respective full MNLI dev sets, while SNLI seeds are comparable in difficulty to their SNLI test problems.  
This difference is statistically significant for both MNLI-m/-mm variant datasets\footnote{A paired t-test comparing model S-acc on Seed vs. Test/Dev was not significant for SNLI ($p=0.6222$), but was significant for both MNLI-m/-mm ($p<0.0001$).
}. A random-subset analysis further supports this result: MNLI-m/-mm seeds were consistently easier than 100 equally sized random subsets drawn from their corresponding Dev sets.\footnote {In comparison, models achieved higher S-acc in 55\% of the randomly drawn SNLI subsamples. The MNLI-m/-mm variants subselected for LLM evaluation are also less difficult, as they were chosen to match the main Seeds in difficulty.}

Overall, our results confirm but are stronger than previous studies that showed models fail to generalize to datasets containing minimal changes \cite{verma-etal-2023-evaluating, arakelyan-etal-2024-semantic, glockner-etal-2018-breaking}, with our work additionally demonstrating that this is true on even easier datasets\footnote{We evaluated the hypothesis-only baselines on the generated variants for SNLI and MNLI and observed that the hypothesis-only artifacts are still substantially present in the variants, with 57\% and 51\% accuracy scores, respectively, in contrast to 33\% of a random baseline.}, obtained from \genmethod. 
Our use of SV-acc especially shows how considering variant proportions per seed can reveal generalization shortcomings, while depending on just sample accuracy S-acc on variants, as is common, often blurs this effect.

\subsection{Which word classes are harder to generalize?}
\begin{figure*}[!t]
  \centering
  \includegraphics[width=\linewidth, trim=0 0 0 00, clip]{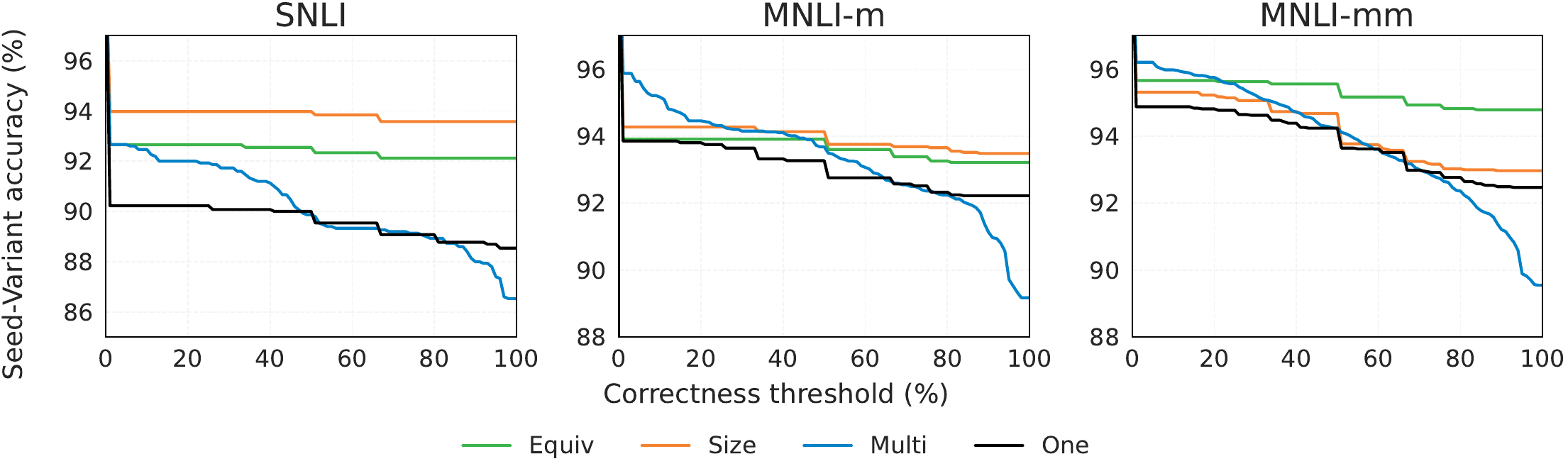}
  \caption{SV curves for ALBERT-XXL on the seed problems that had variants formed with replacements from BERT, RoBERTa, ELECTRA and ALBERT, all sizes tested, grouped by their origin MLM.}
  \label{fig:origin_data_albert}
\end{figure*}
To investigate whether replacing nouns, verbs, or adjectives affects model performance, we compared scores on variants grouped by the open class replaced. We restrict this test to seed problems that share at least two of the three open classes replaced\footnote{For SNLI (N${\text{V}}$=202; N${\text{Adj}}$=130; V${\text{Adj}}$=46); MNLI-m (N${\text{V}}$=422; N${\text{Adj}}$=312; V${\text{Adj}}$=62); and MNLI-mm (N${\text{V}}$=516; N${\text{Adj}}$=360; V${\text{Adj}}$=88).}, such as `A girl carries a small girl', to directly observe the effect of replacing one of the classes as opposed to the other, in this case the noun `girl' opposed to the adjective `small', and separate that from the differences in problem difficulty and imbalanced number of variants across open-classes, as shown in \autoref{section4.table2}.

We found that variants generated by replacing verbs are more difficult than those generated by replacing nouns, followed by adjectives, as shown in the results of the model with the highest averaged S-acc for SNLI and MNLI, ALBERT-XXL,  from \autoref{nvnequal}. In the legend, baseline letters indicate which open class was replaced in seed problems sharing the two open classes in each dataset name. Across datasets, adjectives are initially similar to verbs in lower CTs but become easier in higher ones (green lines). At lower CTs, verbs are easier than nouns in MNLI-m and SNLI, and more difficult than nouns in MNLI-mm (red lines). At high CTs, verbs are more difficult than nouns in MNLI, with nouns being more difficult in SNLI. Lastly, adjectives are easier than nouns in high and low CT, as shown by the blue lines.

\subsection{Do source MLMs affect generalization?}

We further investigated whether NLI models generalize better on variants generated by an MLM with a similar backbone, as it has been suggested in the case of synthetic evaluation datasets that they can favor evaluated models that share the same architecture as the models used to generate the data \cite{li-etal-2020-linguistically, gardner-etal-2020-evaluating}. 

We restricted the set of seed problems to those for which variants could be generated using replacements from each of the following MLMs: BERT, RoBERTa, ELECTRA, and ALBERT.\footnote {Amounting to 150 problems in SNLI, 530 in MNLI-m, and 424 in MNLI-mm, subsampled to isolate the effect of the origin of replacements from that of imbalanced seed problems (see Appendix \autoref{tab:source_models_statistics} for statistics).} We also consider only the results of the corresponding NLI models. 

\autoref{fig:origin_data_albert} shows the SV curves of the best-performing NLI model. Dataset names indicate the MLM used to generate replacements and its relation to the evaluated NLI model: an MLM with the same backbone and size (Equiv); an MLM with the same backbone, but a different size (Size), e.g. BERT-B-S evaluated on BERT-L-generated variants; multiple MLMs, potentially including one with the same backbone as the evaluated NLI model (Multi); or one MLM model with a different architecture (One).\footnote{Note that the same variants, e.g. from BERT-B, can be the equivalent size variants for BERT-B-S, and of different size for BERT-L-S.}

While scores are consistently high on Equiv variants across datasets (green lines), overall scores on other datasets suggest that the MLMs do not favor NLI models that share their architecture.  If they did, we would expect Size to always have the second-highest scores across datasets (orange lines), which is contradicted by its highest scores for SNLI and MNLI-m. Similarly, we would expect One variants to be classified most poorly (black line), followed by 
Multi variants (blue line), since it also includes replacements from MLMs that share the architecture with the evaluated NLI model, which could favor performance. However, the graphs show that models score lowest at higher CTs on Multi.

\subsection{What do the hard NLI problems look like?}
\label{5.5}

We analyzed how successfully, on average, the models classified the original seed problems for all three datasets. We found that all NLI models incorrectly classified 0.7\% of SNLI (16), 0.05\% of MNLI-m (2), and 0.29\% of MNLI-mm (12) problems\footnote{These problems are not in the sub-sample used for LLMs.}. Out of these problems, 66\% had high label variation (see the problems in \autoref{tab:hardets_nli_problems}). For example, a problem with \textit{p} `A man in a colorful shirt and a lady in a white blouse sign copies of books for people' and \textit{h} `Two people sign copies of their latest novel' has been classified as entailment by 3 out of 5 annotators. The world knowledge that people most often sign books they wrote explains the entailment label; however, omitting the adjective `latest' from $h$ also supports the neutral label. This adds evidence to the well-known issue of annotation variation in NLI \cite{pavlick-kwiatkowski-2019-inherent,weber-genzel-etal-2024-varierr}, and higher model error rates on such problems \cite{madaan2024lostinferencerediscoveringrole}. Models also have higher accuracy on variants of seeds they initially classified correctly, as shown by \citet{ohmer-etal-2024-form}: 97\% of their variants are correctly classified, compared to $\approx$ 80\% for seeds which were incorrectly classified. Occasionally, models correctly predicted variants of incorrect seeds (avg. seeds across models of SNLI=98, MNLI-m=207, and MNLI-mm=210).

\section{Conclusions}\label{conclusion} 

We introduced \genmethod{}, a methodology for automatically generating minimal generalization variants of NLI problems designed to preserve reasoning and word overlap. Its strength lies in minimal, controlled replacements with quality filters for fluency and consistency, with \gentest{} being the full methodology of testing models under the simplest generalization conditions.

We show that while NLI models, across sizes and architectures, perform well on variants considered individually, they perform worse when correctness is required for more than about half of the variants of SNLI and MNLI-m/-mm. More specifically, MC values are overall around $\approx$50\%, indicating seed accuracy holds only until that CT. Our results also indicate that variants formed from verbs are generally harder to generalize to than those formed from nouns and adjectives, and that replacements from different MLMs do not affect model scores, suggesting a loss in generalizability regardless of the (mis)match between the origins of MLMs and NLI models. 

Although we currently applied \gentest{} to NLI, we believe that the method is general and robust enough to be extended to other natural language understanding tasks that allow (consistent) replacements of words without affecting the ground truth answers, such as, extractive QA, multiple-choice commonsense/reasoning QA, and semantic textual similarity.

\newpage

\section*{Limitations}\label{limitations}
In \genmethod\ we did not replace words tokenized into subwords, as aggregating their probabilities could introduce artifacts, which could be explored in future work. Additionally, when original words appear multiple times in a problem, we mask only one instance at a time, leaving the others unmasked, which could bias MLM replacements. However, problems with such cases are a small proportion of the seed datasets (SNLI$\approx$2\%, MNLI-m$\approx$6\%, and MNLI-mm$\approx$9\%). Another methodological limitation is that we do not replace morphological derivations of original words to avoid incorrectly replacing them when they appear in nominal compounds. However, this can result in low-fluency variants. For example, not replacing \textit{sleep} from `sleeping' in the hypothesis `She sleeps, and she will be forever sleeping', could result in implausible variants such as `She runs, and she will be forever sleeping'. Such cases are rare in the seed datasets as well (SNLI$\approx$5\%; MNLI-m$\approx$7\%, and MNLI-mm$\approx$6\%). Lastly, our analysis could be extended to include larger LLMs.

\section*{Acknowledgments}
We thank the anonymous reviewers for their useful comments. We also thank the NLP Reading Group at Utrecht University and the NALOMA audience for their useful feedback on this work. 

\bibliography{anthology, custom}

\begin{thebibliography}{45}
\providecommand{\natexlab}[1]{#1}

\bibitem[{Abzianidze et~al.(2023)Abzianidze, Zwarts, and Winter}]{abzianidze-etal-2023-spacenli}
Lasha Abzianidze, Joost Zwarts, and Yoad Winter. 2023.
\newblock \href {https://aclanthology.org/2023.naloma-1.2/} {{S}pace{NLI}: Evaluating the consistency of predicting inferences in space}.
\newblock In \emph{Proceedings of the 4th Natural Logic Meets Machine Learning Workshop}, pages 12--24, Nancy, France. Association for Computational Linguistics.

\bibitem[{Arakelyan et~al.(2024)Arakelyan, Liu, and Augenstein}]{arakelyan-etal-2024-semantic}
Erik Arakelyan, Zhaoqi Liu, and Isabelle Augenstein. 2024.
\newblock \href {https://aclanthology.org/2024.eacl-long.27/} {Semantic sensitivities and inconsistent predictions: Measuring the fragility of {NLI} models}.
\newblock In \emph{Proceedings of the 18th Conference of the European Chapter of the Association for Computational Linguistics (Volume 1: Long Papers)}, pages 432--444, St. Julian{'}s, Malta. Association for Computational Linguistics.

\bibitem[{Bhagavatula et~al.(2019)Bhagavatula, Bras, Malaviya, Sakaguchi, Holtzman, Rashkin, Downey, Yih, and Choi}]{bhagavatula2019abductive}
Chandra Bhagavatula, Ronan~Le Bras, Chaitanya Malaviya, Keisuke Sakaguchi, Ari Holtzman, Hannah Rashkin, Doug Downey, Scott Wen-tau Yih, and Yejin Choi. 2019.
\newblock \href {https://openreview.net/forum?id=Byg1v1HKDB} {Abductive commonsense reasoning}.
\newblock \emph{International Conference on Learning Representations (ICLR)}.

\bibitem[{Bowman et~al.(2015)Bowman, Angeli, Potts, and Manning}]{bowman2015large}
Samuel~R. Bowman, Gabor Angeli, Christopher Potts, and Christopher~D. Manning. 2015.
\newblock \href {https://doi.org/10.18653/v1/D15-1075} {A large annotated corpus for learning natural language inference}.
\newblock In \emph{Proceedings of the 2015 Conference on Empirical Methods in Natural Language Processing}, pages 632--642, Lisbon, Portugal. Association for Computational Linguistics.

\bibitem[{Bowman and Dahl(2021)}]{bowman-dahl-2021-will}
Samuel~R. Bowman and George Dahl. 2021.
\newblock \href {https://doi.org/10.18653/v1/2021.naacl-main.385} {What will it take to fix benchmarking in natural language understanding?}
\newblock In \emph{Proceedings of the 2021 Conference of the North American Chapter of the Association for Computational Linguistics: Human Language Technologies}, pages 4843--4855, Online. Association for Computational Linguistics.

\bibitem[{Budnikov et~al.(2025)Budnikov, Bykova, and Yamshchikov}]{budnikov2025generalization}
Mikhail Budnikov, Anna Bykova, and Ivan~P Yamshchikov. 2025.
\newblock Generalization potential of large language models.
\newblock \emph{Neural Computing and Applications}, 37(4):1973--1997.

\bibitem[{Clark et~al.(2020)Clark, Luong, Le, and Manning}]{clark2020electrapretrainingtextencoders}
Kevin Clark, Minh-Thang Luong, Quoc~V. Le, and Christopher~D. Manning. 2020.
\newblock \href {https://arxiv.org/abs/2003.10555} {Electra: Pre-training text encoders as discriminators rather than generators}.
\newblock \emph{Preprint}, arXiv:2003.10555.

\bibitem[{Dagan et~al.(2005)Dagan, Glickman, and Magnini}]{dagan2005pascal}
Ido Dagan, Oren Glickman, and Bernardo Magnini. 2005.
\newblock The pascal recognising textual entailment challenge.
\newblock In \emph{Machine learning challenges workshop}, pages 177--190. Springer.

\bibitem[{Devlin et~al.(2018)Devlin, Chang, Lee, and Toutanova}]{DBLP:journals/corr/abs-1810-04805}
Jacob Devlin, Ming{-}Wei Chang, Kenton Lee, and Kristina Toutanova. 2018.
\newblock \href {https://arxiv.org/abs/1810.04805} {{BERT:} pre-training of deep bidirectional transformers for language understanding}.
\newblock \emph{CoRR}, abs/1810.04805.

\bibitem[{Dubey et~al.(2024)Dubey, Jauhri, Pandey, Kadian, Al-Dahle, Letman, Mathur, Schelten, Yang, Fan et~al.}]{dubey2024llama}
Abhimanyu Dubey, Abhinav Jauhri, Abhinav Pandey, Abhishek Kadian, Ahmad Al-Dahle, Aiesha Letman, Akhil Mathur, Alan Schelten, Amy Yang, Angela Fan, et~al. 2024.
\newblock The llama 3 herd of models.
\newblock \emph{arXiv e-prints}, pages arXiv--2407.

\bibitem[{Dutt et~al.(2024)Dutt, Ray~Choudhury, Rao, Rose, and Vydiswaran}]{dutt-etal-2024-investigating}
Ritam Dutt, Sagnik Ray~Choudhury, Varun~Venkat Rao, Carolyn Rose, and V.G.Vinod Vydiswaran. 2024.
\newblock \href {https://doi.org/10.18653/v1/2024.genbench-1.11} {Investigating the generalizability of pretrained language models across multiple dimensions: A case study of {NLI} and {MRC}}.
\newblock In \emph{Proceedings of the 2nd GenBench Workshop on Generalisation (Benchmarking) in NLP}, pages 165--182, Miami, Florida, USA. Association for Computational Linguistics.

\bibitem[{Gardner et~al.(2020)Gardner, Artzi, Basmov, Berant, Bogin, Chen, Dasigi, Dua, Elazar, Gottumukkala, Gupta, Hajishirzi, Ilharco, Khashabi, Lin, Liu, Liu, Mulcaire, Ning, Singh, Smith, Subramanian, Tsarfaty, Wallace, Zhang, and Zhou}]{gardner-etal-2020-evaluating}
Matt Gardner, Yoav Artzi, Victoria Basmov, Jonathan Berant, Ben Bogin, Sihao Chen, Pradeep Dasigi, Dheeru Dua, Yanai Elazar, Ananth Gottumukkala, Nitish Gupta, Hannaneh Hajishirzi, Gabriel Ilharco, Daniel Khashabi, Kevin Lin, Jiangming Liu, Nelson~F. Liu, Phoebe Mulcaire, Qiang Ning, Sameer Singh, Noah~A. Smith, Sanjay Subramanian, Reut Tsarfaty, Eric Wallace, Ally Zhang, and Ben Zhou. 2020.
\newblock \href {https://doi.org/10.18653/v1/2020.findings-emnlp.117} {Evaluating models' local decision boundaries via contrast sets}.
\newblock In \emph{Findings of the Association for Computational Linguistics: EMNLP 2020}, pages 1307--1323, Online. Association for Computational Linguistics.

\bibitem[{Glockner et~al.(2018)Glockner, Shwartz, and Goldberg}]{glockner-etal-2018-breaking}
Max Glockner, Vered Shwartz, and Yoav Goldberg. 2018.
\newblock \href {https://doi.org/10.18653/v1/P18-2103} {Breaking {NLI} systems with sentences that require simple lexical inferences}.
\newblock In \emph{Proceedings of the 56th Annual Meeting of the Association for Computational Linguistics (Volume 2: Short Papers)}, pages 650--655, Melbourne, Australia. Association for Computational Linguistics.

\bibitem[{Gururangan et~al.(2018)Gururangan, Swayamdipta, Levy, Schwartz, Bowman, and Smith}]{gururangan-etal-2018-annotation}
Suchin Gururangan, Swabha Swayamdipta, Omer Levy, Roy Schwartz, Samuel Bowman, and Noah~A. Smith. 2018.
\newblock \href {https://doi.org/10.18653/v1/N18-2017} {Annotation artifacts in natural language inference data}.
\newblock In \emph{Proceedings of the 2018 Conference of the North {A}merican Chapter of the Association for Computational Linguistics: Human Language Technologies, Volume 2 (Short Papers)}, pages 107--112, New Orleans, Louisiana. Association for Computational Linguistics.

\bibitem[{He et~al.(2021)He, Liu, Gao, and Chen}]{he2021debertadecodingenhancedbertdisentangled}
Pengcheng He, Xiaodong Liu, Jianfeng Gao, and Weizhu Chen. 2021.
\newblock \href {https://openreview.net/pdf?id=XPZIaotutsD} {Deberta: Decoding-enhanced bert with disentangled attention}.
\newblock \emph{International Conference on Learning Representations (ICLR)}.

\bibitem[{Hupkes et~al.(2023)Hupkes, Giulianelli, Dankers, Artetxe, Elazar, Pimentel, Christodoulopoulos, Lasri, Saphra, Sinclair et~al.}]{hupkes2023taxonomy}
Dieuwke Hupkes, Mario Giulianelli, Verna Dankers, Mikel Artetxe, Yanai Elazar, Tiago Pimentel, Christos Christodoulopoulos, Karim Lasri, Naomi Saphra, Arabella Sinclair, et~al. 2023.
\newblock A taxonomy and review of generalization research in nlp.
\newblock \emph{Nature Machine Intelligence}, 5(10):1161--1174.

\bibitem[{Kaushik et~al.(2020)Kaushik, Hovy, and Lipton}]{kaushik2020learningdifferencemakesdifference}
Divyansh Kaushik, Eduard Hovy, and Zachary~C. Lipton. 2020.
\newblock \href {https://arxiv.org/abs/1909.12434} {Learning the difference that makes a difference with counterfactually-augmented data}.
\newblock \emph{International Conference on Learning Representations (ICLR)}.

\bibitem[{Lampinen et~al.(2024)Lampinen, Dasgupta, Chan, Sheahan, Creswell, Kumaran, McClelland, and Hill}]{dasgupta2024languagemodelshumanlikecontent}
Andrew~K Lampinen, Ishita Dasgupta, Stephanie C~Y Chan, Hannah~R Sheahan, Antonia Creswell, Dharshan Kumaran, James~L McClelland, and Felix Hill. 2024.
\newblock \href {https://doi.org/10.1093/pnasnexus/pgae233} {Language models, like humans, show content effects on reasoning tasks}.
\newblock \emph{PNAS Nexus}, 3(7):pgae233.

\bibitem[{Lan et~al.(2019)Lan, Chen, Goodman, Gimpel, Sharma, and Soricut}]{DBLP:journals/corr/abs-1909-11942}
Zhenzhong Lan, Mingda Chen, Sebastian Goodman, Kevin Gimpel, Piyush Sharma, and Radu Soricut. 2019.
\newblock \href {https://arxiv.org/abs/1909.11942} {{ALBERT:} {A} lite {BERT} for self-supervised learning of language representations}.
\newblock \emph{CoRR}, abs/1909.11942.

\bibitem[{Lewis et~al.(2020)Lewis, Liu, Goyal, Ghazvininejad, Mohamed, Levy, Stoyanov, and Zettlemoyer}]{lewis2019bartdenoisingsequencetosequencepretraining}
Mike Lewis, Yinhan Liu, Naman Goyal, Marjan Ghazvininejad, Abdelrahman Mohamed, Omer Levy, Veselin Stoyanov, and Luke Zettlemoyer. 2020.
\newblock \href {https://doi.org/10.18653/v1/2020.acl-main.703} {{BART}: Denoising sequence-to-sequence pre-training for natural language generation, translation, and comprehension}.
\newblock In \emph{Proceedings of the 58th Annual Meeting of the Association for Computational Linguistics}, pages 7871--7880, Online. Association for Computational Linguistics.

\bibitem[{Li et~al.(2020)Li, Shengshuo, Liu, Wu, Zhou, and Steinert-Threlkeld}]{li-etal-2020-linguistically}
Chuanrong Li, Lin Shengshuo, Zeyu Liu, Xinyi Wu, Xuhui Zhou, and Shane Steinert-Threlkeld. 2020.
\newblock \href {https://doi.org/10.18653/v1/2020.blackboxnlp-1.12} {Linguistically-informed transformations ({LIT}): A method for automatically generating contrast sets}.
\newblock In \emph{Proceedings of the Third BlackboxNLP Workshop on Analyzing and Interpreting Neural Networks for NLP}, pages 126--135, Online. Association for Computational Linguistics.

\bibitem[{Liu et~al.(2019)Liu, Ott, Goyal, Du, Joshi, Chen, Levy, Lewis, Zettlemoyer, and Stoyanov}]{liu2019robertarobustlyoptimizedbert}
Yinhan Liu, Myle Ott, Naman Goyal, Jingfei Du, Mandar Joshi, Danqi Chen, Omer Levy, Mike Lewis, Luke Zettlemoyer, and Veselin Stoyanov. 2019.
\newblock \href {https://arxiv.org/abs/1907.11692} {Roberta: A robustly optimized bert pretraining approach}.
\newblock \emph{Preprint}, arXiv:1907.11692.

\bibitem[{Madaan et~al.(2025)Madaan, Esiobu, Stenetorp, Plank, and Hupkes}]{madaan2024lostinferencerediscoveringrole}
Lovish Madaan, David Esiobu, Pontus Stenetorp, Barbara Plank, and Dieuwke Hupkes. 2025.
\newblock \href {https://doi.org/10.18653/v1/2025.naacl-long.466} {Lost in inference: Rediscovering the role of natural language inference for large language models}.
\newblock In \emph{Proceedings of the 2025 Conference of the Nations of the Americas Chapter of the Association for Computational Linguistics: Human Language Technologies (Volume 1: Long Papers)}, pages 9229--9242, Albuquerque, New Mexico. Association for Computational Linguistics.

\bibitem[{Naik et~al.(2018)Naik, Ravichander, Sadeh, Rose, and Neubig}]{naik-etal-2018-stress}
Aakanksha Naik, Abhilasha Ravichander, Norman Sadeh, Carolyn Rose, and Graham Neubig. 2018.
\newblock \href {https://aclanthology.org/C18-1198/} {Stress test evaluation for natural language inference}.
\newblock In \emph{Proceedings of the 27th International Conference on Computational Linguistics}, pages 2340--2353, Santa Fe, New Mexico, USA. Association for Computational Linguistics.

\bibitem[{Nie et~al.(2019)Nie, Chen, and Bansal}]{nie2019combining}
Yixin Nie, Haonan Chen, and Mohit Bansal. 2019.
\newblock Combining fact extraction and verification with neural semantic matching networks.
\newblock In \emph{Association for the Advancement of Artificial Intelligence ({AAAI})}.

\bibitem[{Nie et~al.(2020)Nie, Williams, Dinan, Bansal, Weston, and Kiela}]{nie2019adversarial}
Yixin Nie, Adina Williams, Emily Dinan, Mohit Bansal, Jason Weston, and Douwe Kiela. 2020.
\newblock \href {https://doi.org/10.18653/v1/2020.acl-main.441} {Adversarial {NLI}: A new benchmark for natural language understanding}.
\newblock In \emph{Proceedings of the 58th Annual Meeting of the Association for Computational Linguistics}, pages 4885--4901, Online. Association for Computational Linguistics.

\bibitem[{Ohmer et~al.(2024)Ohmer, Bruni, and Hupke}]{ohmer-etal-2024-form}
Xenia Ohmer, Elia Bruni, and Dieuwke Hupke. 2024.
\newblock \href {https://doi.org/10.1162/coli_a_00529} {From form(s) to meaning: Probing the semantic depths of language models using multisense consistency}.
\newblock \emph{Computational Linguistics}, 50(3):1507--1556.

\bibitem[{Pavlick and Kwiatkowski(2019)}]{pavlick-kwiatkowski-2019-inherent}
Ellie Pavlick and Tom Kwiatkowski. 2019.
\newblock \href {https://doi.org/10.1162/tacl_a_00293} {Inherent disagreements in human textual inferences}.
\newblock \emph{Transactions of the Association for Computational Linguistics}, 7:677--694.

\bibitem[{Petrov(2025)}]{petrov2025superficial}
Daniel Petrov. 2025.
\newblock \href {https://arxiv.org/abs/2501.02683} {From superficial patterns to semantic understanding: Fine-tuning language models on contrast sets}.
\newblock \emph{arXiv preprint arXiv:2501.02683}.

\bibitem[{Poliak et~al.(2018)Poliak, Naradowsky, Haldar, Rudinger, and Van~Durme}]{poliak-etal-2018-hypothesis}
Adam Poliak, Jason Naradowsky, Aparajita Haldar, Rachel Rudinger, and Benjamin Van~Durme. 2018.
\newblock \href {https://doi.org/10.18653/v1/S18-2023} {Hypothesis only baselines in natural language inference}.
\newblock In \emph{Proceedings of the Seventh Joint Conference on Lexical and Computational Semantics}, pages 180--191, New Orleans, Louisiana. Association for Computational Linguistics.

\bibitem[{Radford et~al.(2019)Radford, Wu, Child, Luan, Amodei, Sutskever et~al.}]{radford2019language}
Alec Radford, Jeffrey Wu, Rewon Child, David Luan, Dario Amodei, Ilya Sutskever, et~al. 2019.
\newblock Language models are unsupervised multitask learners.
\newblock \emph{OpenAI blog}, 1(8):9.

\bibitem[{Rajaee et~al.(2022)Rajaee, Yaghoobzadeh, and Pilehvar}]{rajaee-etal-2022-looking}
Sara Rajaee, Yadollah Yaghoobzadeh, and Mohammad~Taher Pilehvar. 2022.
\newblock \href {https://doi.org/10.18653/v1/2022.emnlp-main.725} {Looking at the overlooked: An analysis on the word-overlap bias in natural language inference}.
\newblock In \emph{Proceedings of the 2022 Conference on Empirical Methods in Natural Language Processing}, pages 10605--10616, Abu Dhabi, United Arab Emirates. Association for Computational Linguistics.

\bibitem[{Rudinger et~al.(2020)Rudinger, Shwartz, Hwang, Bhagavatula, Forbes, Le~Bras, Smith, and Choi}]{rudinger-etal-2020-thinking}
Rachel Rudinger, Vered Shwartz, Jena~D. Hwang, Chandra Bhagavatula, Maxwell Forbes, Ronan Le~Bras, Noah~A. Smith, and Yejin Choi. 2020.
\newblock \href {https://doi.org/10.18653/v1/2020.findings-emnlp.418} {Thinking like a skeptic: Defeasible inference in natural language}.
\newblock In \emph{Findings of the Association for Computational Linguistics: EMNLP 2020}, pages 4661--4675, Online. Association for Computational Linguistics.

\bibitem[{Singh et~al.(2026)Singh, Fry, Perelman, Tart, Ganesh, El-Kishky, McLaughlin, Low, Ostrow, Ananthram, and et~al.}]{singh2026openaigpt5card}
Aaditya Singh, Adam Fry, Adam Perelman, Adam Tart, Adi Ganesh, Ahmed El-Kishky, Aidan McLaughlin, Aiden Low, AJ~Ostrow, Akhila Ananthram, and et~al. 2026.
\newblock \href {https://arxiv.org/abs/2601.03267} {Openai gpt-5 system card}.
\newblock \emph{Preprint}, arXiv:2601.03267.

\bibitem[{Speer(2022)}]{robyn_speer_2022_7199437}
Robyn Speer. 2022.
\newblock \href {https://doi.org/10.5281/zenodo.7199437} {rspeer/wordfreq: v3.0}.

\bibitem[{Srikanth et~al.(2024)Srikanth, Carpuat, and Rudinger}]{srikanth-etal-2024-often}
Neha Srikanth, Marine Carpuat, and Rachel Rudinger. 2024.
\newblock \href {https://doi.org/10.1162/tacl_a_00692} {How often are errors in natural language reasoning due to paraphrastic variability?}
\newblock \emph{Transactions of the Association for Computational Linguistics}, 12:1143--1162.

\bibitem[{Srikanth and Rudinger(2025)}]{srikanth2025nlimicroscopeatomichypothesis}
Neha Srikanth and Rachel Rudinger. 2025.
\newblock \href {https://doi.org/10.18653/v1/2025.naacl-long.130} {{NLI} under the microscope: What atomic hypothesis decomposition reveals}.
\newblock In \emph{Proceedings of the 2025 Conference of the Nations of the Americas Chapter of the Association for Computational Linguistics: Human Language Technologies (Volume 1: Long Papers)}, pages 2574--2589, Albuquerque, New Mexico. Association for Computational Linguistics.

\bibitem[{Team(2024)}]{gemma_2024}
Gemma Team. 2024.
\newblock \href {https://arxiv.org/abs/2408.00118} {Gemma 2: Improving open language models at a practical size}.
\newblock \emph{Preprint}, arXiv:2408.00118.

\bibitem[{Tsuchiya(2018)}]{tsuchiya-2018-performance}
Masatoshi Tsuchiya. 2018.
\newblock \href {https://aclanthology.org/L18-1239/} {Performance impact caused by hidden bias of training data for recognizing textual entailment}.
\newblock In \emph{Proceedings of the Eleventh International Conference on Language Resources and Evaluation ({LREC} 2018)}, Miyazaki, Japan. European Language Resources Association (ELRA).

\bibitem[{Verma et~al.(2023)Verma, Lal, Sinha, Van~Durme, and Poliak}]{verma-etal-2023-evaluating}
Dhruv Verma, Yash~Kumar Lal, Shreyashee Sinha, Benjamin Van~Durme, and Adam Poliak. 2023.
\newblock \href {https://doi.org/10.18653/v1/2023.acl-short.76} {Evaluating paraphrastic robustness in textual entailment models}.
\newblock In \emph{Proceedings of the 61st Annual Meeting of the Association for Computational Linguistics (Volume 2: Short Papers)}, pages 880--892, Toronto, Canada. Association for Computational Linguistics.

\bibitem[{Weber-Genzel et~al.(2024)Weber-Genzel, Peng, De~Marneffe, and Plank}]{weber-genzel-etal-2024-varierr}
Leon Weber-Genzel, Siyao Peng, Marie-Catherine De~Marneffe, and Barbara Plank. 2024.
\newblock \href {https://doi.org/10.18653/v1/2024.acl-long.123} {{V}ari{E}rr {NLI}: Separating annotation error from human label variation}.
\newblock In \emph{Proceedings of the 62nd Annual Meeting of the Association for Computational Linguistics (Volume 1: Long Papers)}, pages 2256--2269, Bangkok, Thailand. Association for Computational Linguistics.

\bibitem[{Williams et~al.(2018)Williams, Nangia, and Bowman}]{williams-etal-2018-broad}
Adina Williams, Nikita Nangia, and Samuel Bowman. 2018.
\newblock \href {https://doi.org/10.18653/v1/N18-1101} {A broad-coverage challenge corpus for sentence understanding through inference}.
\newblock In \emph{Proceedings of the 2018 Conference of the North {A}merican Chapter of the Association for Computational Linguistics: Human Language Technologies, Volume 1 (Long Papers)}, pages 1112--1122, New Orleans, Louisiana. Association for Computational Linguistics.

\bibitem[{Yang et~al.(2023)Yang, Song, Ren, Lyu, Wang, Zhuo, Liu, Wang, Foster, and Zhang}]{yang-etal-2023-distribution}
Linyi Yang, Yaoxian Song, Xuan Ren, Chenyang Lyu, Yidong Wang, Jingming Zhuo, Lingqiao Liu, Jindong Wang, Jennifer Foster, and Yue Zhang. 2023.
\newblock \href {https://doi.org/10.18653/v1/2023.emnlp-main.276} {Out-of-distribution generalization in natural language processing: Past, present, and future}.
\newblock In \emph{Proceedings of the 2023 Conference on Empirical Methods in Natural Language Processing}, pages 4533--4559, Singapore. Association for Computational Linguistics.

\bibitem[{Yang et~al.(2019)Yang, Dai, Yang, Carbonell, Salakhutdinov, and Le}]{DBLP:journals/corr/abs-1906-08237}
Zhilin Yang, Zihang Dai, Yiming Yang, Jaime~G. Carbonell, Ruslan Salakhutdinov, and Quoc~V. Le. 2019.
\newblock \href {https://arxiv.org/abs/1906.08237} {Xlnet: Generalized autoregressive pretraining for language understanding}.
\newblock \emph{CoRR}, abs/1906.08237.

\bibitem[{Zhang et~al.(2022)Zhang, Roller, Goyal, Artetxe, Chen, Chen, Dewan, Diab, Li, Lin, Mihaylov, Ott, Shleifer, Shuster, Simig, Koura, Sridhar, Wang, and Zettlemoyer}]{zhang2022opt}
Susan Zhang, Stephen Roller, Naman Goyal, Mikel Artetxe, Moya Chen, Shuohui Chen, Christopher Dewan, Mona Diab, Xian Li, Xi~Victoria Lin, Todor Mihaylov, Myle Ott, Sam Shleifer, Kurt Shuster, Daniel Simig, Punit~Singh Koura, Anjali Sridhar, Tianlu Wang, and Luke Zettlemoyer. 2022.
\newblock \href {https://arxiv.org/abs/2205.01068} {Opt: Open pre-trained transformer language models}.
\newblock \emph{Preprint}, arXiv:2205.01068.

\end{thebibliography}

\appendix

\section{Appendix}
\label{sec:appendix}
\subsection{Variant Creation}

\begin{table}[h]
\centering
\begin{tabularx}{.95\linewidth}{
>{\raggedright\arraybackslash}p{1.6cm}
>{\centering\arraybackslash}X
>{\centering\arraybackslash}X
>{\centering\arraybackslash}X
>{\centering\arraybackslash}X
>{\centering\arraybackslash}X}
\textbf{Dataset} & \textbf{R} & \textbf{\#PS} & \textbf{N(\%)} & \textbf{C(\%)} & \textbf{E(\%)} \\
\midrule
\multirow{4}{*}{SNLI}
 & N & 7363 & 33 & 29 & 38 \\
 & V & 3780 & 32 & 28 & 40 \\
 & Adj & 1067 & 34 & 24 & 42 \\
 & Adv & 76 & 26 & 13 & 61 \\
\midrule
\multirow{3}{*}{MNLI-m}
 & N & 5979 & 29 & 31 & 40 \\
 & V & 4438 & 27 & 32 & 41 \\
& Adj & 1923 & 26 & 25 & 48 \\
\midrule
\multirow{3}{*}{MNLI-mm}
 & N & 6640 & 29 & 31 & 40 \\
 & V & 4614 & 27 & 32 & 42 \\
 & Adj & 2197 & 26 & 28 & 47 \\
\bottomrule
\end{tabularx}
\vspace{-2mm}
\caption{Number of potential seed problems from SNLI-test and the development set of MNLI-m/-mm (\#PS). Columns R, N, C, and E show the divisions by replaced open-class and gold label (\% neutral, contradiction, entailment). We ultimately excluded adverbs because of their final low seed count in SNLI.}
\label{tab:label_counts_with_total}
\end{table}

\begin{table}[t]
\centering
\scriptsize
\setlength{\tabcolsep}{3pt}
\begin{tabular}{p{3.5cm}ccc}
\toprule
\textbf{Model} & \textbf{Size} & \textbf{Architecture} & \textbf{Vocabulary} \\
\midrule
BERT \cite{DBLP:journals/corr/abs-1810-04805} & B & E & 28,996 \\
\textcolor{gray}{\tiny google-bert/bert-base-cased} & & & \\[0.3em]

BERT  & L & E & 28,996 \\
\textcolor{gray}{\tiny google-bert/bert-large-cased} & & & \\[0.3em]

RoBERTa \cite{liu2019robertarobustlyoptimizedbert} & B & E & 50,265 \\
\textcolor{gray}{\tiny FacebookAI/roberta-base} & & & \\[0.3em]

RoBERTa & L & E & 50,265 \\
\textcolor{gray}{\tiny FacebookAI/roberta-large} & & & \\[0.3em]

BART \cite{lewis2019bartdenoisingsequencetosequencepretraining} & B & E-D & 50,265 \\
\textcolor{gray}{\tiny facebook/bart-base} & & & \\[0.3em]

BART & L & E-D & 50,265 \\
\textcolor{gray}{\tiny facebook/bart-large} & & & \\[0.3em]

ALBERT \cite{DBLP:journals/corr/abs-1909-11942} & B & E & 30,000 \\
\textcolor{gray}{\tiny albert/albert-base-v2} & & & \\[0.3em]

ALBERT & XXL & E & 30,000 \\
\textcolor{gray}{\tiny albert/albert-xxlarge-v2} & & & \\[0.3em]

ELECTRA \cite{clark2020electrapretrainingtextencoders} & B & E & 30,522 \\
\textcolor{gray}{\tiny google/electra-base-generator} & & & \\[0.3em]

ELECTRA & L & E & 30,522 \\
\textcolor{gray}{\tiny google/electra-large-generator} & & & \\

\bottomrule
\end{tabular}
\caption{Overview of MLMs used for replacement generation. The Size, Architecture, and Vocabulary columns show the model size (Base -- B, Large -- L, or XXLarge -- XXL), architecture (Encoder, Decoder, or Encoder-Decoder), and vocabulary size.}
\label{tab:_MLM_overview}
\end{table}
\paragraph{MLMs} The MLMs used for replacement generation are shown in \autoref{tab:_MLM_overview}. In addition, we considered DeBERTa, but its replacements were frequently too implausible to include. From the models shown in \autoref{tab:_MLM_overview}, we excluded BART because it contributed more to bad variants in SNLI, as explained in \autoref{annotation_gudidelines}. We did not consider BART by default for MNLI-m/-mm replacements,  to avoid creating plausibility differences across variants from different datasets.
\begin{table*}[!t]
\centering
\setlength{\tabcolsep}{3.5pt}
\renewcommand{\arraystretch}{0.88}

\begin{tabular}{@{}lcc*{10}{r}@{}}
\toprule
D & C & M 
& A-B & A-XXL 
& BA-B & BA-L 
& BE-B & BE-L 
& E-B & E-L 
& R-B & R-L \\
\midrule

\multirow{6}{*}{SNLI}
& A & POS  & 52 & 52 & 36 & 35 & 52 & 53 & 52 & 52 & 49 & 50 \\
&   & Prob & 17 & 18 & 15 & 16 & 10 & 8  & 14 & 12 & 9  & 8  \\
& N & POS  & 93 & 94 & 78 & 75 & 91 & 91 & 90 & 90 & 92 & 92 \\
&   & Prob & 18 & 20 & 14 & 16 & 13 & 11 & 16 & 14 & 10 & 8  \\
& V & POS  & 72 & 76 & 63 & 63 & 72 & 71 & 72 & 74 & 68 & 68 \\
&   & Prob & 12 & 13 & 12 & 13 & 9  & 7  & 12 & 11 & 7  & 6  \\

\midrule

\multirow{6}{*}{MNLI-m}
& A & POS  & 53 & 54 & 37 & 34 & 53 & 53 & 53 & 53 & 53 & 53 \\
&   & Prob & 34 & 22 & 28 & 27 & 19 & 15 & 22 & 20 & 18 & 15 \\
& N & POS  & 88 & 90 & 73 & 72 & 86 & 86 & 86 & 86 & 88 & 88 \\
&   & Prob & 34 & 23 & 28 & 26 & 22 & 18 & 26 & 22 & 19 & 16 \\
& V & POS  & 75 & 76 & 58 & 59 & 73 & 73 & 71 & 71 & 71 & 71 \\
&   & Prob & 11 & 8  & 11 & 11 & 7  & 6  & 8  & 7  & 6  & 5  \\

\midrule

\multirow{6}{*}{MNLI-mm}
& A & POS  & 55 & 56 & 38 & 34 & 56 & 56 & 54 & 55 & 54 & 55 \\
&   & Prob & 33 & 22 & 27 & 28 & 18 & 16 & 22 & 19 & 17 & 14 \\
& N & POS  & 89 & 90 & 73 & 72 & 87 & 88 & 86 & 87 & 89 & 89 \\
&   & Prob & 32 & 21 & 27 & 26 & 21 & 18 & 25 & 21 & 19 & 16 \\
& V & POS  & 74 & 75 & 58 & 60 & 72 & 72 & 70 & 70 & 71 & 71 \\
&   & Prob& 11 & 7  & 11 & 10 & 6  & 5  & 8  & 6  & 6  & 5  \\

\bottomrule

\end{tabular}

\caption{The average percentage of $r$ with the same open class (POS) and equal or higher probability than $o$ (Prob) for sentences ($p$ and $h$) across the three datasets. Column names indicate the origin model of $r$, e.g., ALBERT (A), BART (BA), BERT (BE), ELECTRA (E), and RoBERTa (R), for different model sizes.}
\label{avrgredc}
\end{table*}

\paragraph{Class \& Probability Filtering} 
\begin{figure*}[!b]
  \centering
  \includegraphics[width=\textwidth, trim=00 00 0 00, clip]{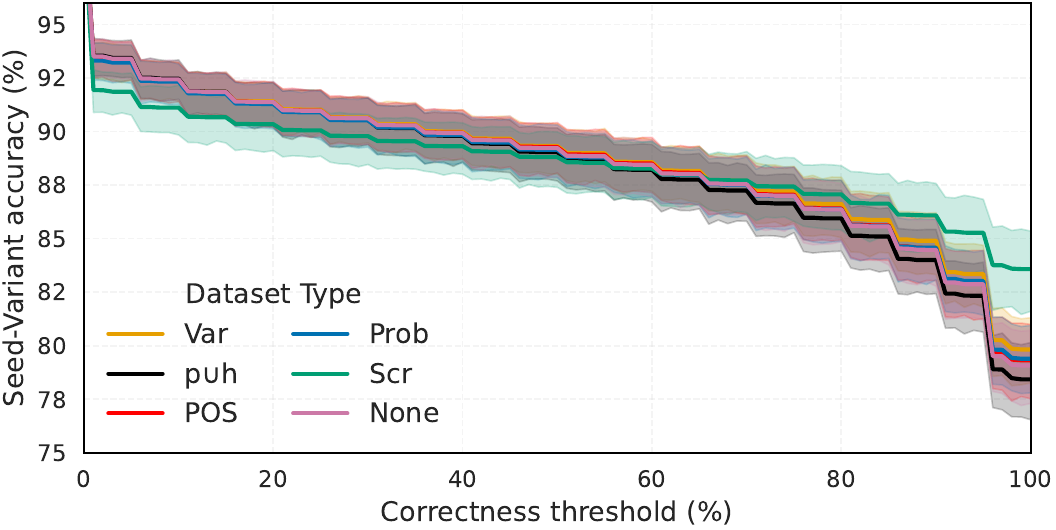} 
  \caption{Averaged SV curves of all models on SNLI datasets formed with different quality filtering criteria, for all CTs.}
  \label{fig:tests_noisy_full}
\end{figure*}
We POS-tagged replacements $r$ in the context of their corresponding problems using the spaCy model
\texttt{en\_core\_web\_sm}, and excluded those whose word classes differed from the original word $o$. We also recorded the probability the source MLM assigned to each $r$ of $o$ in $p$ or $h$, and excluded those with a lower probability than the original word. \genmethod\ uses probability as a safeguarding criterion, assuming that replacements as likely as $o$ are less likely to be semantically implausible. Original words $o_i$ that were not part of the model's vocabulary were also automatically excluded. Note that both probability and open-class filtering are important, as highly probable replacements might not have the correct open-class and vice versa. \autoref{avrgredc} shows the percentages of replacements with higher probability than $o$ under rows Prob, and with the correct open-class in POS.

\subsection{Which filtering criteria matter?}\label{filtering_criteria}

To test if different quality filtering criteria for $r$ affect NLI models' scores, we formed new variants for the seed problems of SNLI Var with $r$ i) from the union of $p$ and $h$, instead of their intersection in \autoref{eq:p_h_intersection} -- $_{p \cup h}$; ii) only having $o^c$ -- $_\text{POS}$; iii) only having $o_{i>}$ -- $_\text{Prob}$; iv) of any open-class or probability -- $_\text{None}$; v) with their letters randomly scrambled -- $_\text{Scr}$. We still excluded punctuation signs and $r$ already part of $p$ and $h$. Three out of the five datasets had more unique variants given the less strict filtering criteria, i.e. $\approx$380k, while $_\text{Scr}$ and Var have $\approx$ 190k.
We plot the averaged SV-acc of all models in  \autoref{fig:tests_noisy_full}. 

While $_\text{Scr}$ starts as the lowest curve (in green) in \autoref{fig:tests_noisy_full}, at very high CTs, models end up having the highest robustness on it, surprisingly. This was also shown with the nonce words in \citet{dasgupta2024languagemodelshumanlikecontent}. 

The other variant datasets seem to follow a constant downward trend, namely, the highest score is achieved on Var, followed by $_\text{Prob}$, $_\text{POS}$, $_\text{None}$, and $_{p \cup h}$, where the more unique variants are considered, the lower the scores get. Out of these, $_{p \cup h}$, $_\text{POS}$, and $_\text{None}$ have the lowest scores, indicating that not controlling for probability or plausibility might disadvantage the models. Thus, the higher scores achieved on Var show that our enforced quality criteria successfully create `friendly' tests.

\subsection{Annotation}\label{annotation_gudidelines} Two authors annotated variants for SNLI, and one for MNLI. We evaluated replacements based on how much they changed fluency and the original logical label of the NLI problems, which we considered correct. The scores used were: 1 -- poor, 2 -- mostly poor, 3 -- uncertain, 4 -- mostly good, 5 -- good. \autoref{tab:nli-examples-annotation} shows the annotation guidelines, alongside an explanation for them. Variants were classified as poor if they: 1) had ungrammatical replacements making manual validation difficult; 2) had missing arguments; 3) were nonsense; or 4) were logic-non-preserving. Aspects 1) and 2) were chosen given that they directly hinder the evaluation of fluency and reasoning.

The instructions in the annotation guidelines are provided below with the demo examples in \autoref{tab:nli-examples-annotation}:

\begin{quotation}
\small
The NLI problems are assessed on fluency (grammaticality and sensibility) and reasoning. The reasoning component focuses on the relation between the meaning of p and h rather than their fluency.

The original NLI problems could suffer from fluency and reasoning; however, the obtained NLI problem variants should be assessed with respect to the original NLI problems. Annotation should assess the fluency and reasoning of the variants while assuming that the fluency and reasoning of the original NLI problems are fine. That's why variant NLI problems are provided with the origin word, which helps to reconstruct the original NLI problem.

Below are several examples to demonstrate different combinations of fluency and reasoning. The scale 1-5 should be interpreted as: poor, mostly poor, uncertain, mostly good, good.  
\end{quotation}

\begin{figure}[!t] 
    \centering
    \vspace{2mm}
    \includegraphics[width=\linewidth, trim=0 30 0 45, clip]{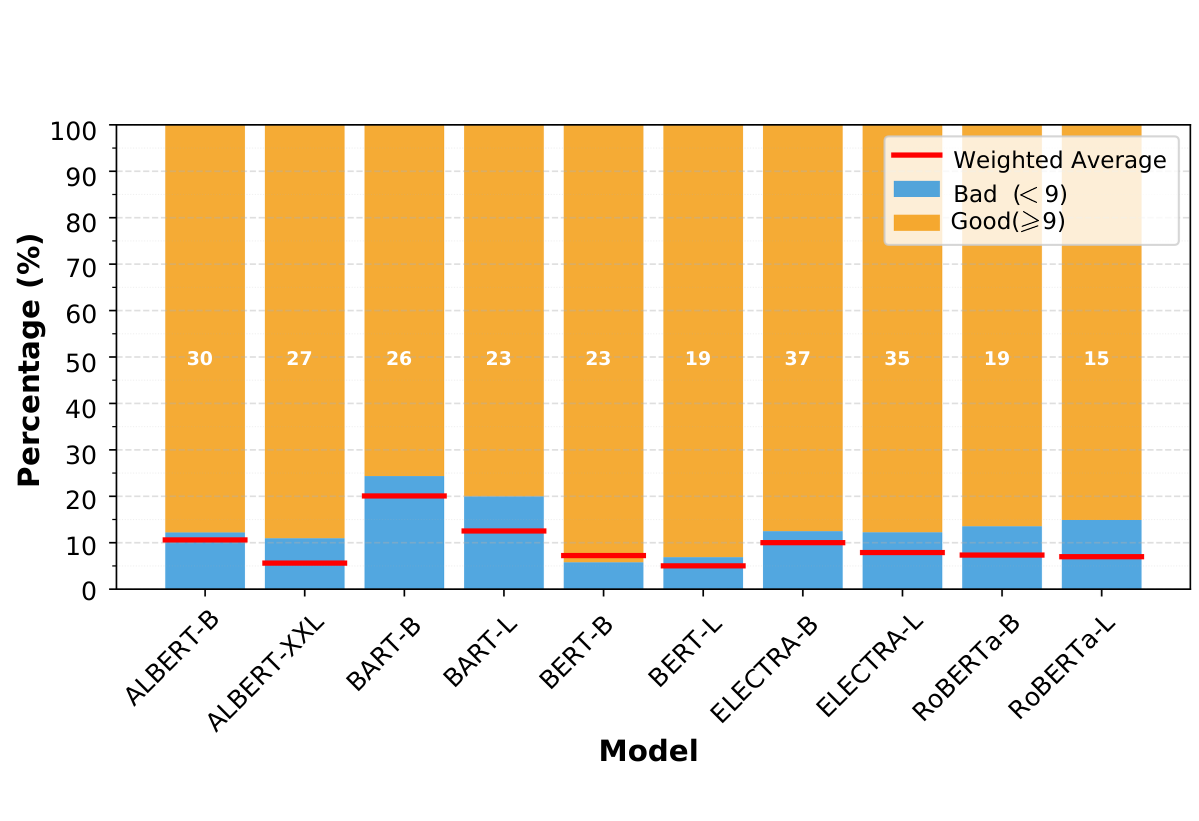}
    \caption{Averaged fluency and reasoning scores for the normalized counts of 100 random SNLI variants for nouns, verbs, and adjectives. The red lines show bar plots weighted by the distribution of open classes in the SNLI seed problems (N=67\%, V=23\%, ADJ=10\%). Good variants have a score of $F+R\geq 9$.}
    \label{fig:combined-threshold-9}
\end{figure}

\begin{figure}[!b]
    \vspace{2mm}
   
    \includegraphics[width=\linewidth, trim=0 00 0 25, clip]{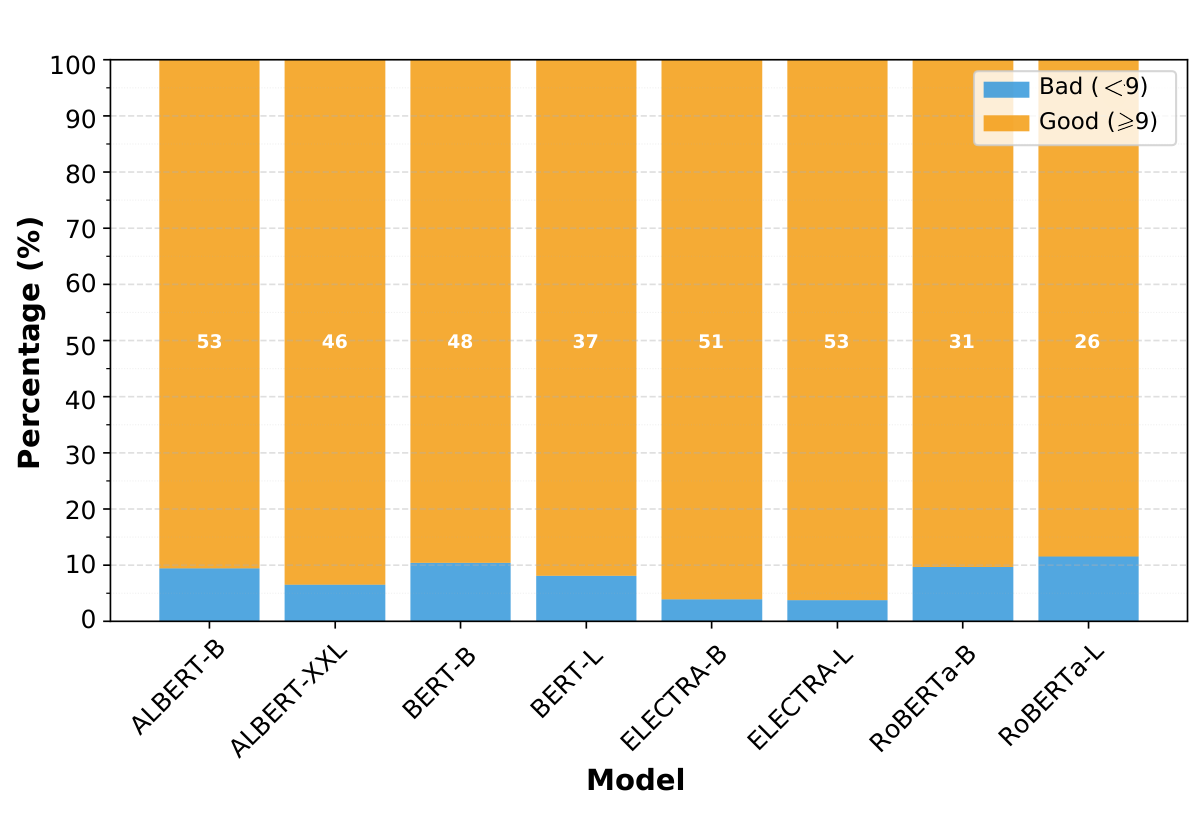}
    \caption{Fluency and reasoning scores for 100 randomly sampled SNLI variants across classes, their normalized counts, and the models that validated them, after exclusion of replacements from BART. Note that 91 out of the 100 examples evaluated had $F+R\geq 9$.}
    \label{fig:AFTER_EVAL_scores}
\end{figure}
\begin{figure}[!t] 
    \centering
    \includegraphics[width=\linewidth, trim=25 30 5 55, clip]{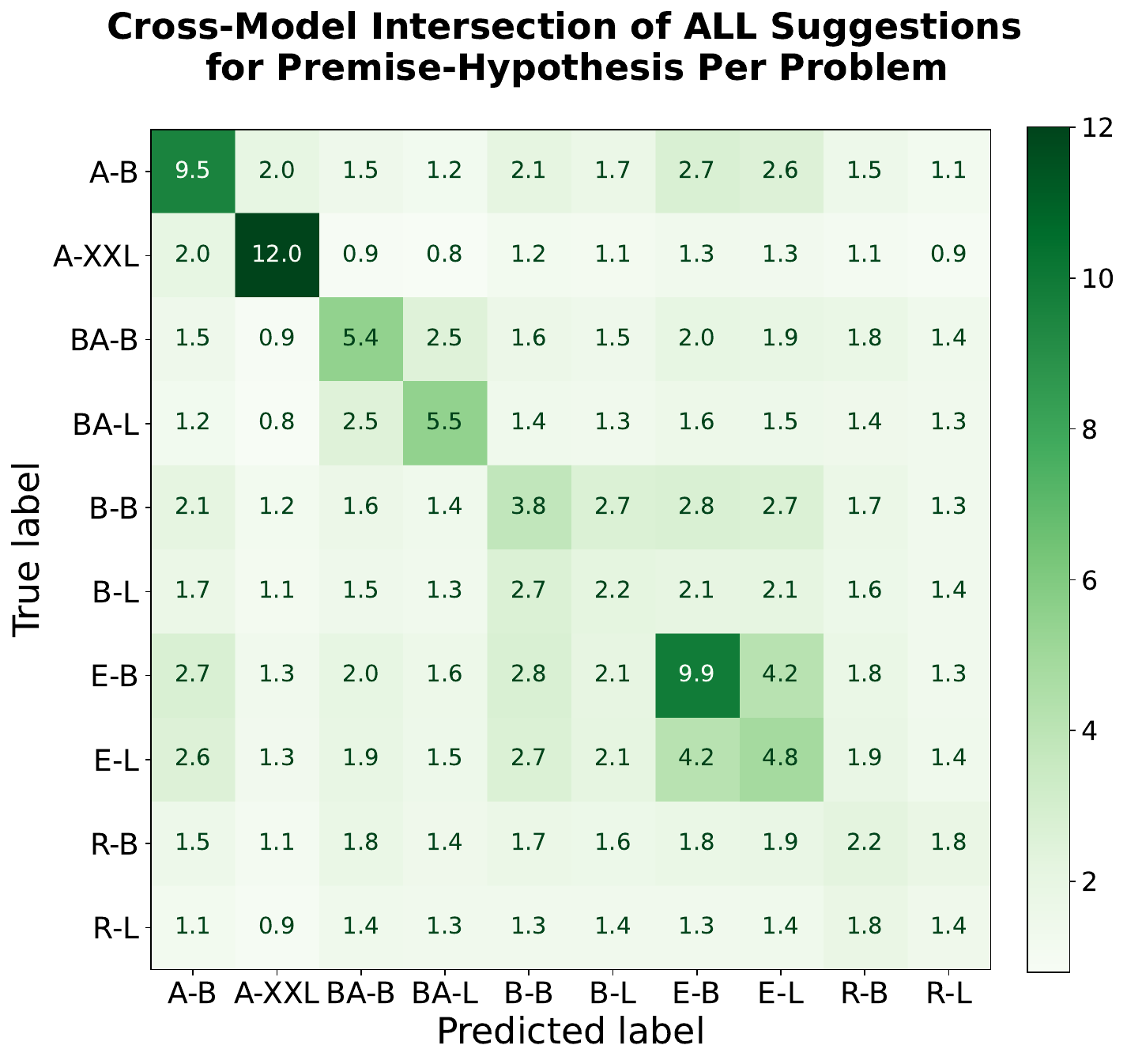}

    \caption{Pairwise overlap matrix showing how models' replacements intersect on average, across problems, with the averaged number of model-specific replacements on the diagonal. For instance, a replacement like `bear' counts on the diagonal if it is unique to one model, or in the other cells if it is shared between models.}
    \label{fig:confusionmatrix}
\end{figure}

In the first stage of annotation, only a few SNLI (5\%) variants received scores lower than 3 on R, due to a lack of label preservation or ungrammaticalities. After this first stage, BART replacements were excluded as they contributed more to bad variants, as shown in \autoref{fig:combined-threshold-9}, resulting in a minor reduction of 318 SNLI seed problems. We did not exclude other models, since the pairwise overlap matrix in \autoref{fig:confusionmatrix} showed more unique than overlapping replacements across models, with only BERT and ELECTRA showing nearly equal proportions of both.  

Given that only one author annotated all MNLI-m/-mm variants (800), we asked GPT-5.3 \cite{singh2026openaigpt5card} to flag implausible or non-preserving reasoning variants. The model flagged 18 examples (15 for plausibility and 3 for reasoning), out of which only 1 was implausible, also identified by the previous author annotator: e.g. $p$: `Well, it would probably been things like living back in the hills of TN and having to ride a mule to wau, things like that’, $h$: `While living in the hills of TN, I rode a mule to wau’, where the word ‘school’ was replaced by ‘wau’.
\begin{table}[!b]
\centering
\resizebox{\columnwidth}{!}{\begin{tabular}{r rr rr rr}
\toprule
 & \multicolumn{2}{c}{\textbf{SNLI}} & \multicolumn{2}{c}{\textbf{MNLI-m}} & \multicolumn{2}{c}{\textbf{MNLI-mm}} \\
\cmidrule(lr){2-3} \cmidrule(lr){4-5} \cmidrule(lr){6-7}
\textbf{Model} & Raw & \% & Raw & \% & Raw & \% \\
\midrule
ALBERT-B    & 7,814   & 5  & 37,568  & 10 & 40,802  & 10 \\
ALBERT-XXL  & 12,820  & 7  & 13,657  & 4  & 15,784  & 4  \\
BERT-B      & 1,979   & 1  & 4,308   & 1  & 5,209   & 1  \\
BERT-L      & 1,043   & 1  & 2,405   & 1  & 2,890   & 1  \\
ELECTRA-B   & 7,750   & 5  & 10,156  & 3  & 11,117  & 3  \\
ELECTRA-L   & 2,827   & 2  & 4,749   & 1  & 5,476   & 1  \\
RoBERTa-B   & 1,441   & 1  & 4,114   & 1  & 4,658   & 1  \\
RoBERTa-L   & 775     & 0  & 2,825   & 1  & 3,242   & 1  \\
Multi        & 135,615 & 79 & 281,866 & 78 & 325,113 & 78 \\
\bottomrule
\end{tabular}}
\caption{Variants across datasets grouped by the MLM that suggested their replacement, with raw and \% counts shown. The 
Multi-row shows replacements suggested by at least two MLMs.}
\label{tab:source_models_statistics}
\end{table}

\begin{table}[!t]
\centering
\small
\begin{tabularx}{\linewidth}{lX}
\textbf{Evaluated Model} & \textbf{Model Card} \\
\midrule

\multicolumn{2}{c}{\textit{SNLI-only models}} \\
BART-B-S & \href{https://huggingface.co/varun-v-rao/bart-base-snli-model1}{varun-v-rao/bart-base-snli-model1} \\
BERT-B-S & \href{https://huggingface.co/textattack/bert-base-uncased-snli}{textattack/bert-base-uncased-snli} \\
BERT-L-S & \href{https://huggingface.co/varun-v-rao/bert-large-cased-lora-1.58M-snli}{varun-v-rao/bert-large-cased-lora-1.58M-snli} \\
DeBERTa-v3-B-S & \href{https://huggingface.co/pepa/deberta-v3-base-snli}{pepa/deberta-v3-base-snli} \\
DeBERTa-v3-L-S & \href{https://huggingface.co/pepa/deberta-v3-large-snli}{pepa/deberta-v3-large-snli} \\
GPT-2-L-S & \href{https://huggingface.co/varun-v-rao/gpt2-large-snli-model3}{varun-v-rao/gpt2-large-snli-model3} \\
OPT-1-3b-S & \href{https://huggingface.co/utahnlp/snli_facebook_opt-1.3b_seed-3}{utahnlp/snli\_facebook\_opt-1.3b\_seed-3} \\
RoBERTa-B-S & \href{https://huggingface.co/pepa/roberta-base-snli}{pepa/roberta-base-snli}\\
\midrule
\multicolumn{2}{c}{\textit{MNLI-only models}} \\
BERT-B-M & \href{https://huggingface.co/textattack/bert-base-uncased-mnli}{textattack/bert-base-uncased-mnli} \\
RoBERTa-B-M & \href{https://huggingface.co/roberta-base-mnli}{roberta-base-mnli} \\
RoBERTa-L-M & \href{https://huggingface.co/roberta-large-mnli}{roberta-large-mnli} \\
BART-L-M & \href{https://huggingface.co/facebook/bart-large-mnli}{facebook/bart-large-mnli} \\
\midrule
\multicolumn{2}{c}{\textit{SMFA models}} \\
ALBERT-XXL-SMFA & \href{https://huggingface.co/ynie/albert-xxlarge-v2-snli_mnli_fever_anli_R1_R2_R3-nli}{ynie/albert-xxlarge-v2-snli\_mnli\_fever\_anli\_R1\_R2\_R3-nli} \\

BART-L-SMFA & \href{https://huggingface.co/ynie/bart-large-snli_mnli_fever_anli_R1_R2_R3-nli}{ynie/bart-large-snli\_mnli\_fever\_anli\_R1\_R2\_R3-nli} \\

ELECTRA-L-SMFA & \href{https://huggingface.co/ynie/electra-large-discriminator-snli_mnli_fever_anli_R1_R2_R3-nli}{ynie/electra-large-discriminator-snli\_mnli\_fever\_anli\_R1\_R2\_R3-nli} \\

RoBERTa-L-SMFA & \href{https://huggingface.co/ynie/roberta-large-snli_mnli_fever_anli_R1_R2_R3-nli}{ynie/roberta-large-snli\_mnli\_fever\_anli\_R1\_R2\_R3-nli} \\
XLNet-L-SMFA & \href{https://huggingface.co/ynie/xlnet-large-cased-snli_mnli_fever_anli_R1_R2_R3-nli}{ynie/xlnet-large-cased-snli\_mnli\_fever\_anli\_R1\_R2\_R3-nli} \\

\midrule
\multicolumn{2}{c}{\textit{LLMs}} \\

Gemma-2-9b-exc/mix & \href{https://huggingface.co/google/gemma-2-9b-it}{google/gemma-2-9b-it} \\
Llama-3.1-8B-exc/mix & \href{https://huggingface.co/meta-llama/Meta-Llama-3.1-8B-Instruct}{meta-llama/Meta-Llama-3.1-8B}  \\

\bottomrule
\end{tabularx}
\caption{Evaluated models and their Hugging Face references. For LLMs, the suffix \textit{-mix/-exc} indicates which type of few-shot examples were used in the prompts, i.e. exclusively from the dataset the variants are based on, or a mix of both SNLI and MNLI datasets.}
\label{tab:appendix-model-cards}
\end{table}

\subsection{Evaluated models}\label{prompting}
\paragraph{Evaluated Models} 

The evaluated models and their model cards are shown in \autoref{tab:appendix-model-cards}.
\begin{table}[!t]
\centering
\footnotesize
\setlength{\tabcolsep}{6pt}
\begin{tabular}{p{0.75\columnwidth}}
\toprule
\textbf{\textbf{Prompt Template} }\\
\midrule
\begin{minipage}[t]{\linewidth}
\ttfamily
\raggedright
\{\% for x in few\_shot -\%\}\\
Premise: \{\{ x["premise"] \}\}\\
Hypothesis: \{\{ x["hypothesis"] \}\}\\
A. Entailment\\
B. Neutral\\
C. Contradiction\\
Answer: \{\{ x["answer"] \}\}\\[4pt]
\{\% endfor -\%\}\\
Premise: \{\{ premise \}\}\\
Hypothesis: \{\{ hypothesis \}\}\\
A. Entailment\\
B. Neutral\\
C. Contradiction\\
Answer: \{\{ choice\_text, e.g. A \}\}
\end{minipage}
\\
\bottomrule
\end{tabular}
\caption{Prompt template used for LLM evaluation from \citet{madaan2024lostinferencerediscoveringrole}.}
\label{tab:prompt_template}
\end{table}
\begin{table}[!t]
\centering
\begin{tabular}{lcc}
\toprule
Comparison & Selected & All \\
\midrule

\multicolumn{3}{l}{\textit{SNLI}} \\
S--SV 50       & 0.0118 & 0.0002 \\
S--SV 60       & 0.0027 & $<0.001$ \\
S--SV 70       & 0.0020 & $<0.001$ \\
S--SV 85       & 0.0037 & $<0.001$ \\
S--Test/Dev         & 0.6222 & 0.9867 \\
\midrule

\multicolumn{3}{l}{\textit{MNLI-m}} \\
S--SV 50       & 0.0776 & 0.0087 \\
S--SV 60       & 0.0003 & $<0.001$ \\
S--SV 70       & 0.0002 & $<0.001$ \\
S--SV 85       & 0.0005 & $<0.001$ \\
S--Test/Dev          & 0.0018 & $<0.001$ \\
\midrule

\multicolumn{3}{l}{\textit{MNLI-mm}} \\
S--SV 50       & 0.1738 & 0.0285 \\
S--SV 60       & 0.0001 & $<0.001$ \\
S--SV 70       & 0.0009 & $<0.001$ \\
S--SV 85       & 0.0017 & $<0.001$ \\
S--Test/Dev          & $<0.001$ & $<0.001$ \\
\bottomrule
\end{tabular}

\caption{P-values from paired $t$-tests comparing Seed S-acc
with SV-acc at different correctness thresholds and with
Test/Dev S-acc. Selected refers to the two best-performing
and one worst-performing model shown in
\autoref{fig:figure_3}; All includes all evaluated models.}
\label{tab:ttest}
\end{table}
\subsubsection{LLMs}
\paragraph{Seed problems} For each seed dataset, we randomly formed 100 subsets of 500 randomly sampled seed problems and compared models' NLI S-acc on them. Model scores on subselected seeds varied by around 3\% from their scores on the bigger seed datasets, indicating that random subsampling roughly preserves the original seed distribution. From these subsets, we selected a random sample from each dataset to test LLMs on.
\paragraph{Instructing and evaluation} We obtained logits for \texttt{google/gemma-2-9b} and \texttt{meta-llama/Llama-3.1-8B} with: temperature=0, max. no. of tokens=20, torch dtype=torch.bfloat16, and attention = sdpa. The prompt structure is shown in \autoref{tab:prompt_template}, while the few-shot examples per dataset appear in \autoref{tab:few_shot_examples}. Note that the few-shot examples are balanced across labels and are of 2 types, i.e. randomly drawn from the variants' source dataset, or a mix from both datasets. With our evaluation method, we replicated the scores of \citet{madaan2024lostinferencerediscoveringrole} for Llama-3.1-8B on SNLI-dev twice: with the current batch of exclusively SNLI-dev shots from \autoref{tab:few_shot_examples}, and 6 other randomly chosen ones from SNLI-test. In both cases, our scores were around 74-76\% accuracy, which is slightly higher than their reported 70\% for 5-shot examples. Finally, for our variant evaluation, we evaluated both LLMs only once over the set of unique variants in $_\text{Var}$ to halve computation time. We investigated whether running Llama directly on the full variant dataset, which contains duplicates, changes its predictions, and found that it does so in 0.1\% of cases. 

\FloatBarrier

\begin{table*}[htbp]
\centering
\renewcommand{\arraystretch}{1.2}
\small
\begin{tabularx}{\textwidth}{p{2.5cm} p{2.5cm} c c c c c X}
\textbf{Premise} & \textbf{Hypothesis} & \textbf{Label} & \textbf{Original} & \textbf{Suggested} & \textbf{F} & \textbf{R} & \textbf{Explanation} \\
\hline
A man in a black shirt, in a detached kitchen, holding up meat he took out of a bag. & A woman in a black shirt, in a detached kitchen, holding up meat he took out of a bag. & C & commercial & detached & 5 & 5 & \textbf{Good}: the variant is fluent and preserves reasoning. \\
A teenage dog is running in a field near a mountain. & A teenage dog is running outdoors. & E & yellow & teenage & 5 & 5 &  \textbf{Good}: the variant is fluent and preserves reasoning. \\
A man driving while at a restaurant. & A man driving while at a restaurant eating. & N & laughing & driving & 4 & 5 & \textbf{Mostly good}: the variant might be likely in a certain scenario (e.g. there might be a restaurant where tables are like cars). \\
A man is celebrating his victory while smiling and wearing champagne in the air with his teammate. & A man is happily celebrating his victory while smiling and wearing champagne in the air & N & shooting & wearing & 3 & 5 & \textbf{Uncertain}: the variant might be likely in a certain scenario (e.g. the man is wearing a champagne-shaped costume), but a less likely one. \\
A man slung into the ear wearing a striped shirt in a small boat filled with many people. & A man is slung into the ear and wearing a light striped shirt. & N & pointing & slung & 2 & 5 & \textbf{Mostly poor}: the scenario is very unlikely, e.g. a man being forced to hear something. \\
clutched to her ear, a woman bends forward at the side of a busy street. & clutched to her ear, a woman bends forward & E & Phone & clutched & 1 & 5 & \textbf{Poor}: the variant is ungrammatical, thus making it difficult to verify its fluency, i.e. missing the theme of `clutched'. \\
A hole is on a cherry picker in a palm tree. & A hole falls out of a tree. & C & worker & hole & 1 & 5 & \textbf{Poor}: the variant is not fluent until we consider a very specific metaphorical context, i.e., a hole cannot have agency. \\
A shirt booth with a man got a shirt. & The man is got some pants. & C & printing & got & 1 & 4 & \textbf{Mostly Good}: the variant is still fluent, despite being metaphorical, and it is still likely to preserve the initial inference label, despite the hypothesis's ungrammaticality. \\
This child is took a pedicure. & Child took a manicure. & C & getting & took & 1 & 3 & \textbf{Uncertain}: the ungrammaticality of the premise makes it difficult to assess its fluency; however, the main source of the contrast giving the contradiction (i.e. \textit{pedicure} vs. \textit{manicure}) is kept. \\
A brown dog wearing a collar is chasing and running on a red broom. & There is an animal running a broom. & E & biting & running & 4 & 2 & \textbf{Very Poor}: Even though `running a broom' might be running \textit{with} a broom, running \textit{on} something does not entail running \textit{with} X. \\
 Two construction workers produced the steel ribbed exterior of a new building at their work site. & Two workers are produced a building & N & climbing & produced & 1 & 1 & \textbf{Poor}: the hypothesis is hard to understand given its ungrammaticality, while the subject of `produced' is unclear, which makes reasoning impossible to assess.
\end{tabularx}
\caption{Examples of annotation scores for NLI variants, considering their fluency (F) and the preservation of the original NLI label (R), alongside explanations for their scores. Note that the labels under Column Label are neutral (N), entailment (E), and contradiction (C).}
\label{tab:nli-examples-annotation}
\end{table*}

\begin{table*}[p]
\centering
\footnotesize
\setlength{\tabcolsep}{6pt}
\begin{tabular}{p{0.15\textwidth} p{0.75\textwidth}}
\toprule
\textbf{Dataset} & \textbf{Few-shot examples} \\
\midrule
SNLI &
\begin{minipage}[t]{0.75\textwidth}
\ttfamily\scriptsize
Premise: A man with a beard skateboarding and a boy with a blue and black backpack riding a green bike in the background.\\
Hypothesis: There is a man and a boy outside.\\
A. Entailment \quad B. Neutral \quad C. Contradiction \quad Answer: A\\[3pt]
Premise: There is a room full of pictures all in the wall and a woman in a coat is looking back over her shoulders strangely.\\
Hypothesis: the room is large\\
A. Entailment \quad B. Neutral \quad C. Contradiction \quad Answer: B\\[3pt]
Premise: A Black woman on the street is talking on her cellphone.\\
Hypothesis: A black woman is on the payphone, ordering pizza\\
A. Entailment \quad B. Neutral \quad C. Contradiction \quad Answer: C\\[3pt]
Premise: The side of a building next to a church is painted with a brightly colored Coca-Cola sign.\\
Hypothesis: There is a Coca-Cola sign on a church next to a building.\\
A. Entailment \quad B. Neutral \quad C. Contradiction \quad Answer: C\\[3pt]
Premise: A butterfly costumed girl waves at the crowd.\\
Hypothesis: A butterfly costumed girl is standing on the stage.\\
A. Entailment \quad B. Neutral \quad C. Contradiction \quad Answer: B\\[3pt]
Premise: A man holding his two adorable babies.\\
Hypothesis: A man has more than one child.\\
A. Entailment \quad B. Neutral \quad C. Contradiction \quad Answer: A
\end{minipage}
\\
\midrule
MNLI &
\begin{minipage}[t]{0.75\textwidth}
\ttfamily\scriptsize
Premise: Ras Mohammed National Park has over 1,500 species of fish and 150 types of coral along with an offshore vertical sea wall.\\
Hypothesis: The national park contains many types of fish and coral.\\
A. Entailment \quad B. Neutral \quad C. Contradiction \quad Answer: A\\[3pt]
Premise: Celebrating Trajan's campaigns against the Dacians in what is now Romania, the minutely detailed friezes spiraling around the column constitute a veritable textbook of Roman warfare utilizing some 2,500 figures.\\
Hypothesis: The friezes were added to the column four centuries ago.\\
A. Entailment \quad B. Neutral \quad C. Contradiction \quad Answer: B\\[3pt]
Premise: And fell.\\
Hypothesis: Did not fall.\\
A. Entailment \quad B. Neutral \quad C. Contradiction \quad Answer: C\\[3pt]
Premise: On the radio, they are still talking about the Brooklyn Museum's controversial art exhibit.\\
Hypothesis: The Brooklyn Museum has an exhibit that is making people talk.\\
A. Entailment \quad B. Neutral \quad C. Contradiction \quad Answer: A\\[3pt]
Premise: and for i don't know it must have been two or three weeks there they were doing this expanded nightly news\\
Hypothesis: The nightly news wanted to provide more extensive coverage of what was happening.\\
A. Entailment \quad B. Neutral \quad C. Contradiction \quad Answer: B\\[3pt]
Premise: The sheets are creamy white and the tissue lining in the envelope a bluer white.\\
Hypothesis: The sheets and the lining are black.\\
A. Entailment \quad B. Neutral \quad C. Contradiction \quad Answer: C
\end{minipage}
\\
\midrule
Mixed &
\begin{minipage}[t]{0.75\textwidth}
\ttfamily\scriptsize
Premise: A man with a beard skateboarding and a boy with a blue and black backpack riding a green bike in the background.\\
Hypothesis: There is a man and a boy outside.\\
A. Entailment \quad B. Neutral \quad C. Contradiction \quad Answer: A\\[3pt]
Premise: There is a room full of pictures all in the wall and a woman in a coat is looking back over her shoulders strangely.\\
Hypothesis: the room is large\\
A. Entailment \quad B. Neutral \quad C. Contradiction \quad Answer: B\\[3pt]
Premise: A Black woman on the street is talking on her cellphone.\\
Hypothesis: A black woman is on the payphone, ordering pizza\\
A. Entailment \quad B. Neutral \quad C. Contradiction \quad Answer: C\\[3pt]
Premise: This was enlarged by Herod, sacked in a.d. 70, and totally flattened by Emperor Hadrian in a.d. 135.\\
Hypothesis: 65 year passed between being sacked and flattened.\\
A. Entailment \quad B. Neutral \quad C. Contradiction \quad Answer: A\\[3pt]
Premise: it's slow it's uh there are many better machines on the market right now for\\
Hypothesis: This machine is too old, that's why it's slow.\\
A. Entailment \quad B. Neutral \quad C. Contradiction \quad Answer: B\\[3pt]
Premise: well i think it made parts of it a lot easier i i is this your first that you're having\\
Hypothesis: I think it did not help at all.\\
A. Entailment \quad B. Neutral \quad C. Contradiction \quad Answer: C
\end{minipage}
\\
\bottomrule
\end{tabular}
\caption{Few-shot examples used for LLM evaluation, containing, when datasets are mentioned separately, random examples exclusively drawn from SNLI-dev for SNLI variants or MNLI-train for MNLI variants, with the mixed row contains 3 examples from each SNLI-dev and MNLI-train.}
\label{tab:few_shot_examples}
\end{table*}
\begin{figure*}[htbp]
  \centering
  \includegraphics[width=\textwidth]{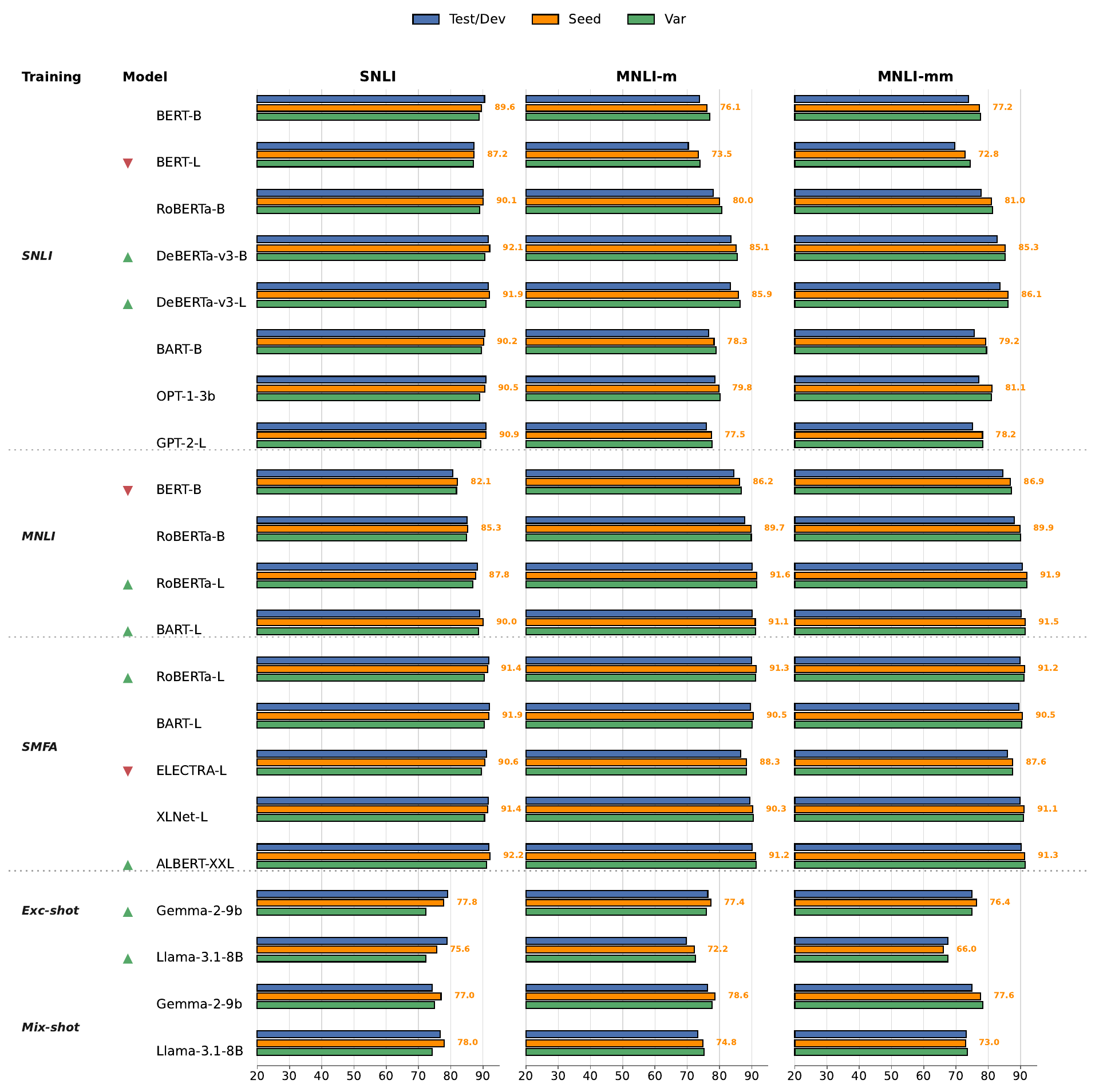} 
  \vspace{4mm}
  \caption{Results of all models, grouped by their training either fine-tuned on SNLI, MNLI, SMFA, or LLMs prompted with examples exclusively chosen from SNLI-dev (for SNLI-var) or MNLI-train (for MNLI-m/-mm). Test/Dev scores are S-acc on the test problems for SNLI and dev problems for MNLI, while Seed and Var are S-acc scores on seed and variant problems. The green triangles in front of the models in each training category show the two best models, while the red ones show the worst.}
  \label{fig:figure_15}
\end{figure*}

\begin{figure*}[t]
    \centering
    \includegraphics[width=\textwidth]{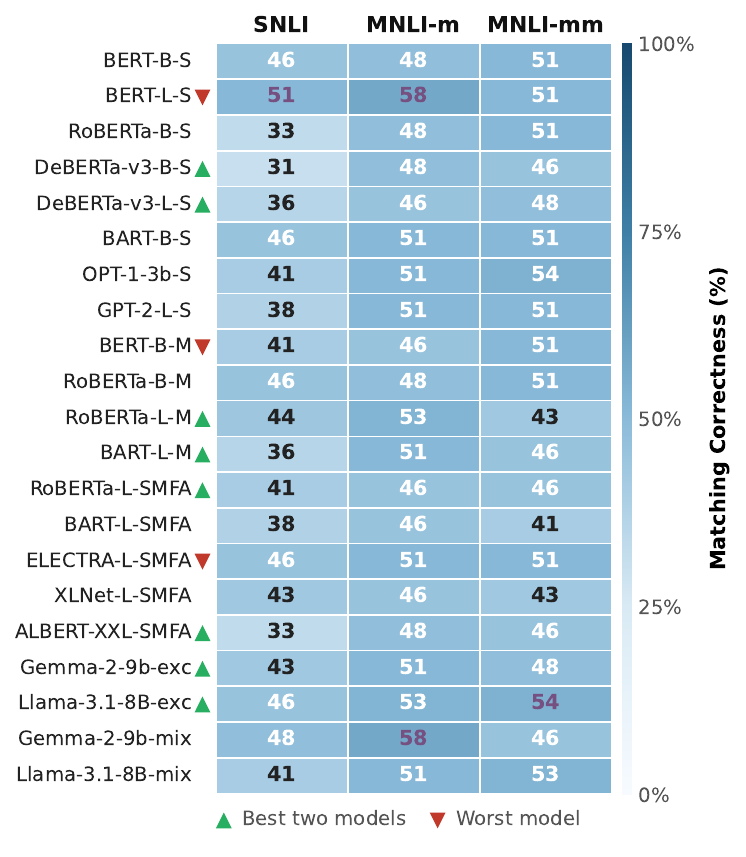}
    \caption{MCs across all models and datasets:  the numbers show on which CT models get the closest SV-acc on variants to their initial S-acc on seed problems.  As most numbers are $\approx$ 50\%, they indicate that models do not generalize to more variants than that. Purple numbers mark the highest MCs per dataset, while black ones indicate MCs under 45.}
    \label{fig:heatmap_all_models}
\end{figure*}

\begin{figure*}[!htbp]
  \centering
  \includegraphics[height=0.85\textheight]{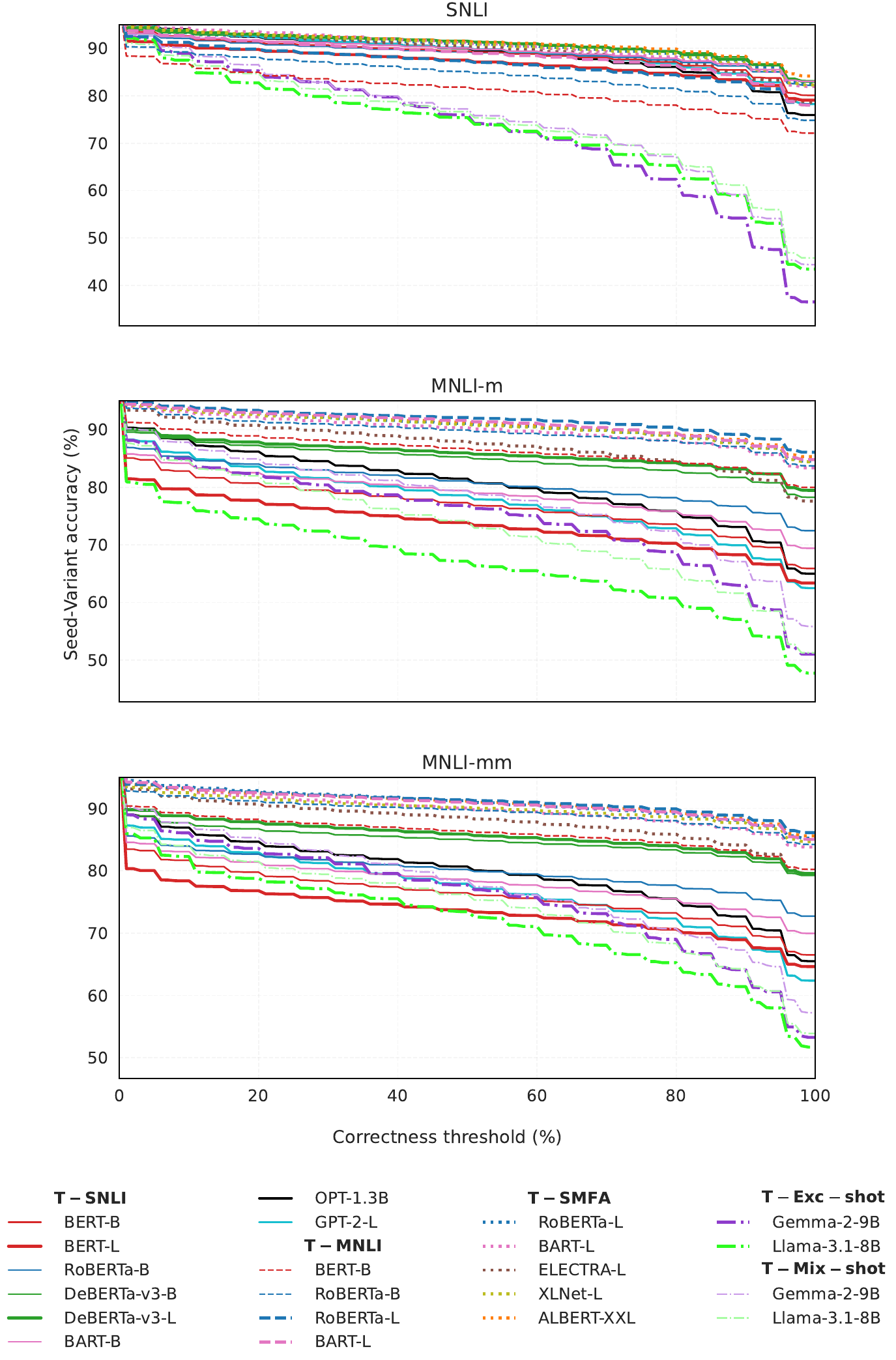} 
  \vspace{4mm}
  \caption{SV-acc curves on variants from SNLI and MNLI-m/-mm. The legend shows NLI models and the datasets they were fine-tuned on, i.e., SNLI, MNLI, or a combination of NLI datasets e.g., SMFA, or the type of few-shot prompts we used for LLMs, i.e., exclusively taken from SNLI-dev (for SNLI variants) or MNLI-train (for MNLI-m/-mm variants) or a mix of both MNLI and SNLI.}
  \label{fig:allposallmodels08}
\end{figure*}

\begin{figure*}[!htbp]
    \centering
    \includegraphics[width=\textwidth]{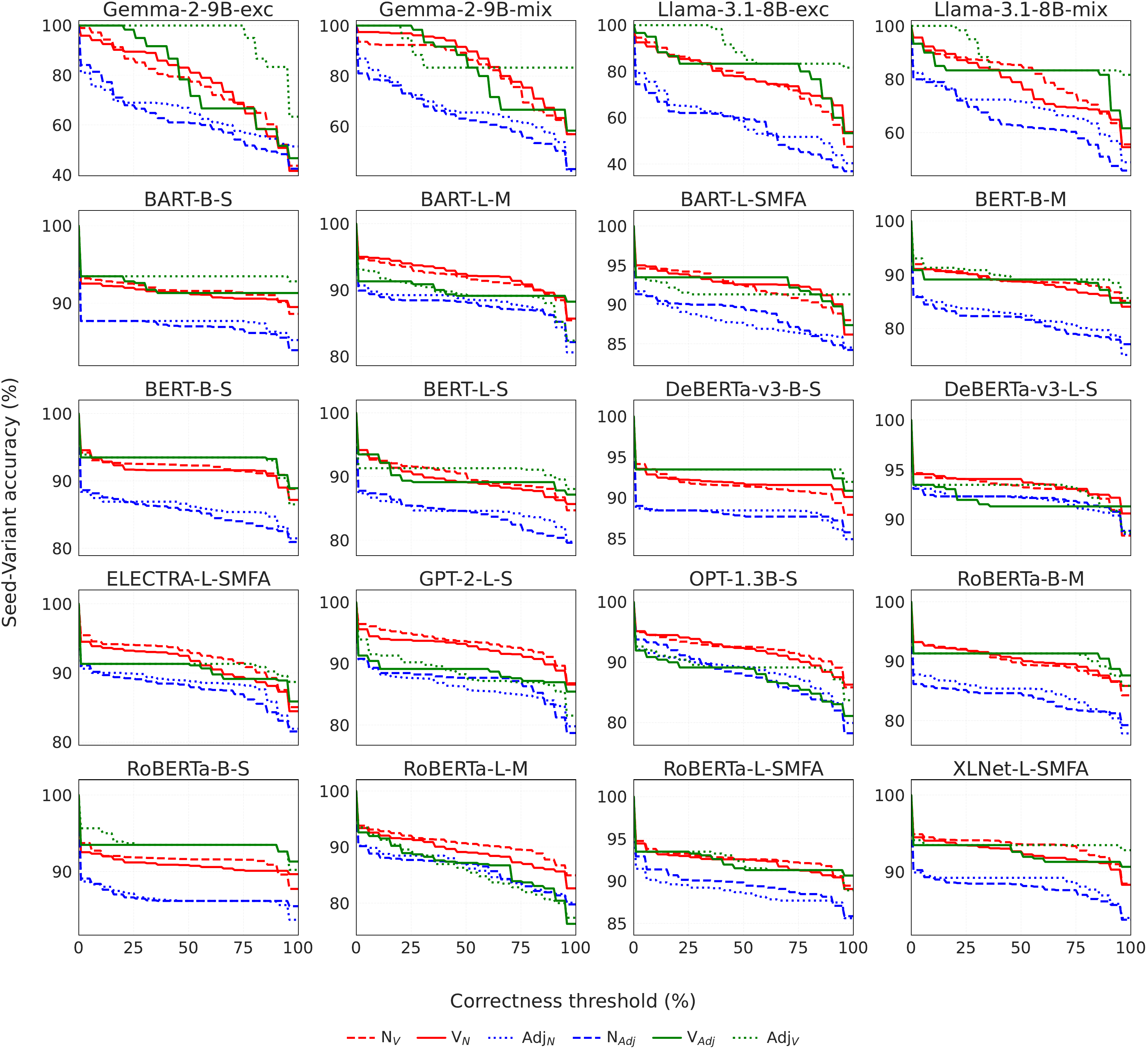}
    \caption{SV-acc for SNLI seed problems with at least two out of the three replaced open classes in each seed (N${\text{V}}$=202; N${\text{Adj}}$=130; V${\text{Adj}}$=46). The legend shows, by the larger letter of each dataset, which open class was replaced in seed problems that shared at least two.}
    \label{fig:figure_nvadj_appendix_base}
\end{figure*}
\begin{figure*}[!htbp]
    \centering
    \includegraphics[width=\textwidth]{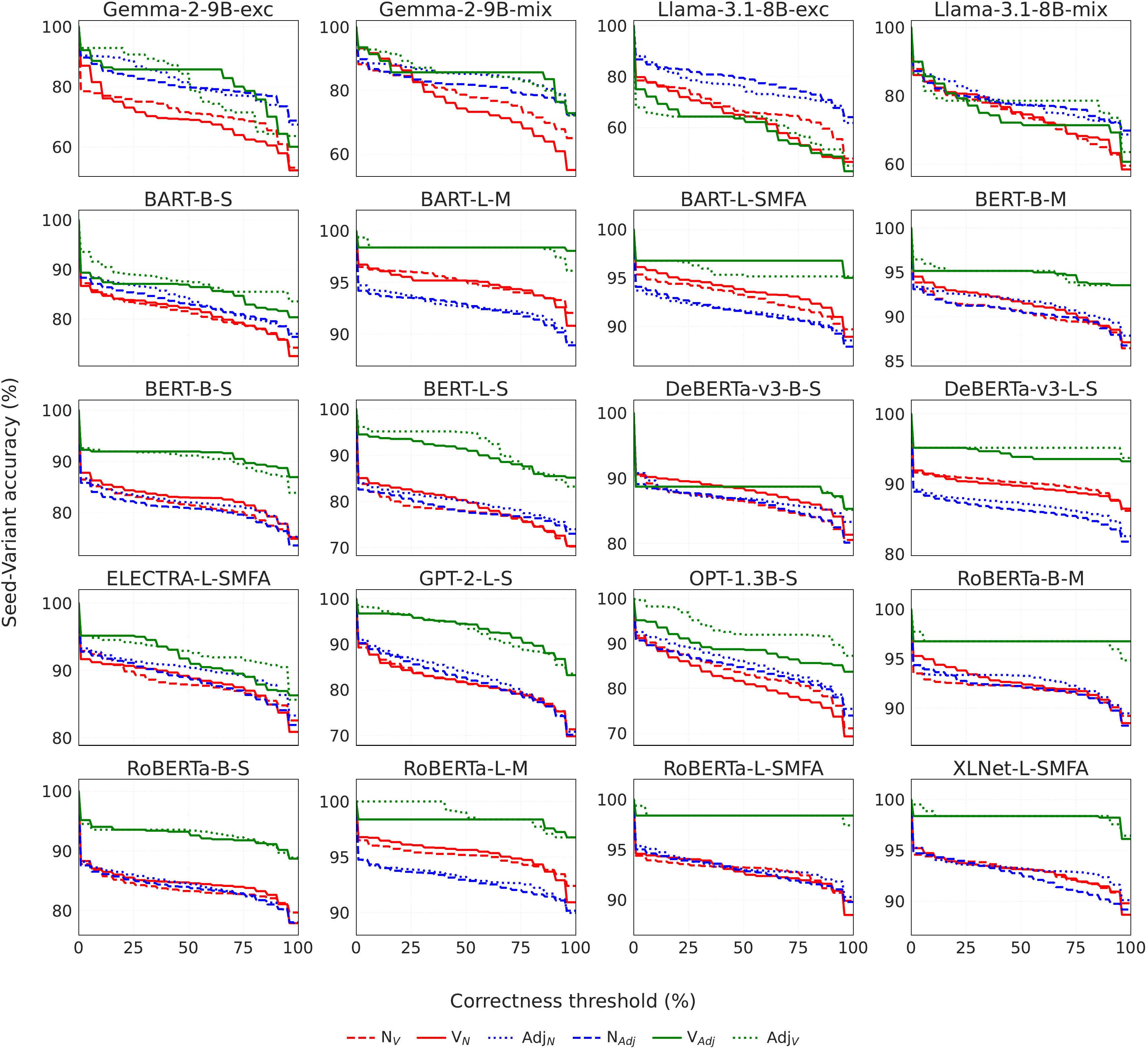}
    \caption{SV-acc for MNLI-mm seed problems with at least two out of the three replaced open classes in the seed problems (N${\text{V}}$=516; N${\text{Adj}}$=360; V${\text{Adj}}$=88).}
    \label{fig:figure_nvadj_appendix_mm}
\end{figure*}
\begin{figure*}[!t]
    \centering
    \includegraphics[width=\textwidth]{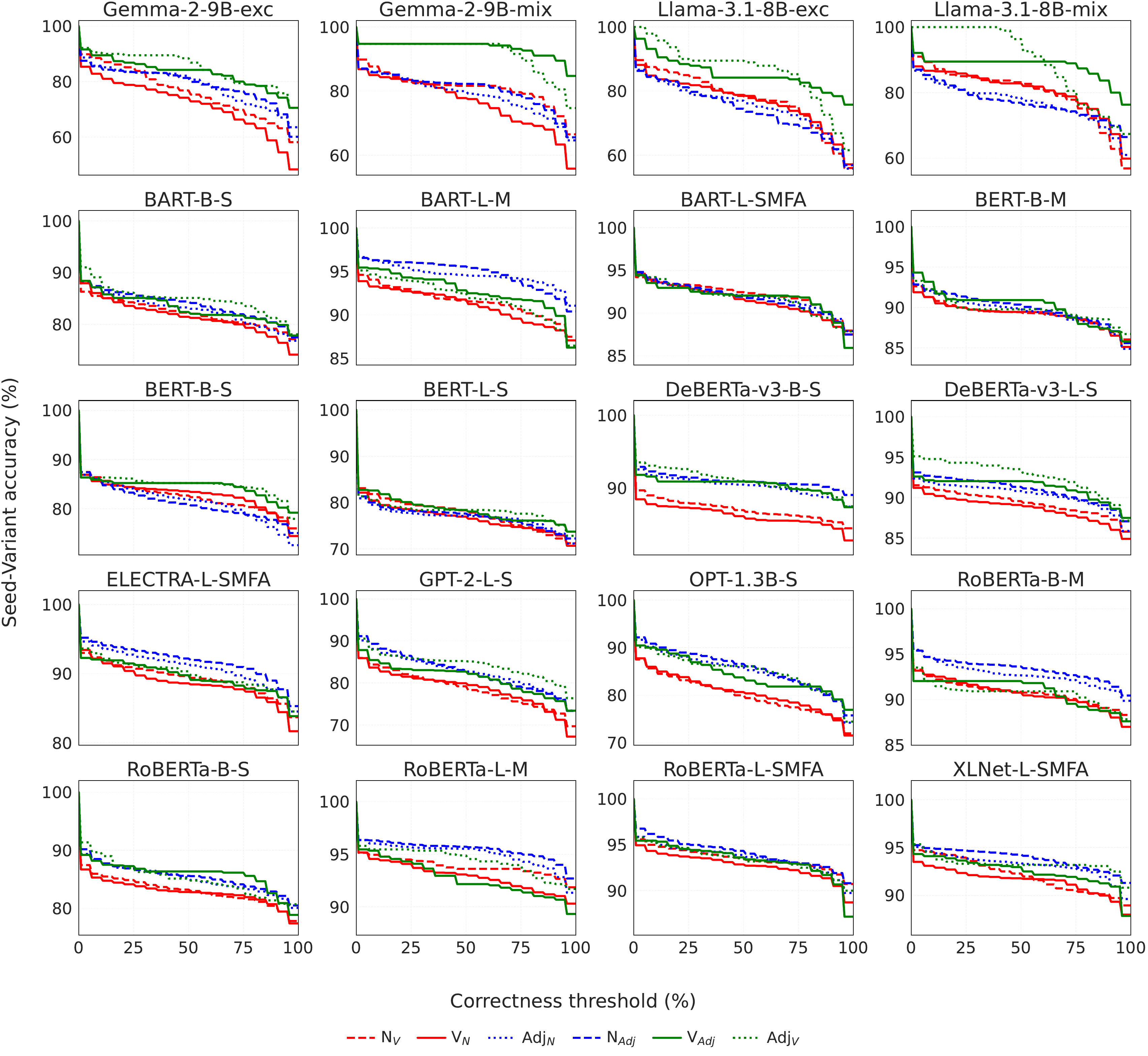}
    \caption{SV-acc for MNLI-m seed problems that have at least two out of the three replaced open classes in them (N${\text{V}}$=422; N${\text{Adj}}$=312; V${\text{Adj}}$=62).}
    \label{fig:figure_nvadj_appendix_m}
\end{figure*}

\begin{figure*}[t]
    \centering
    \includegraphics[width=\textwidth]{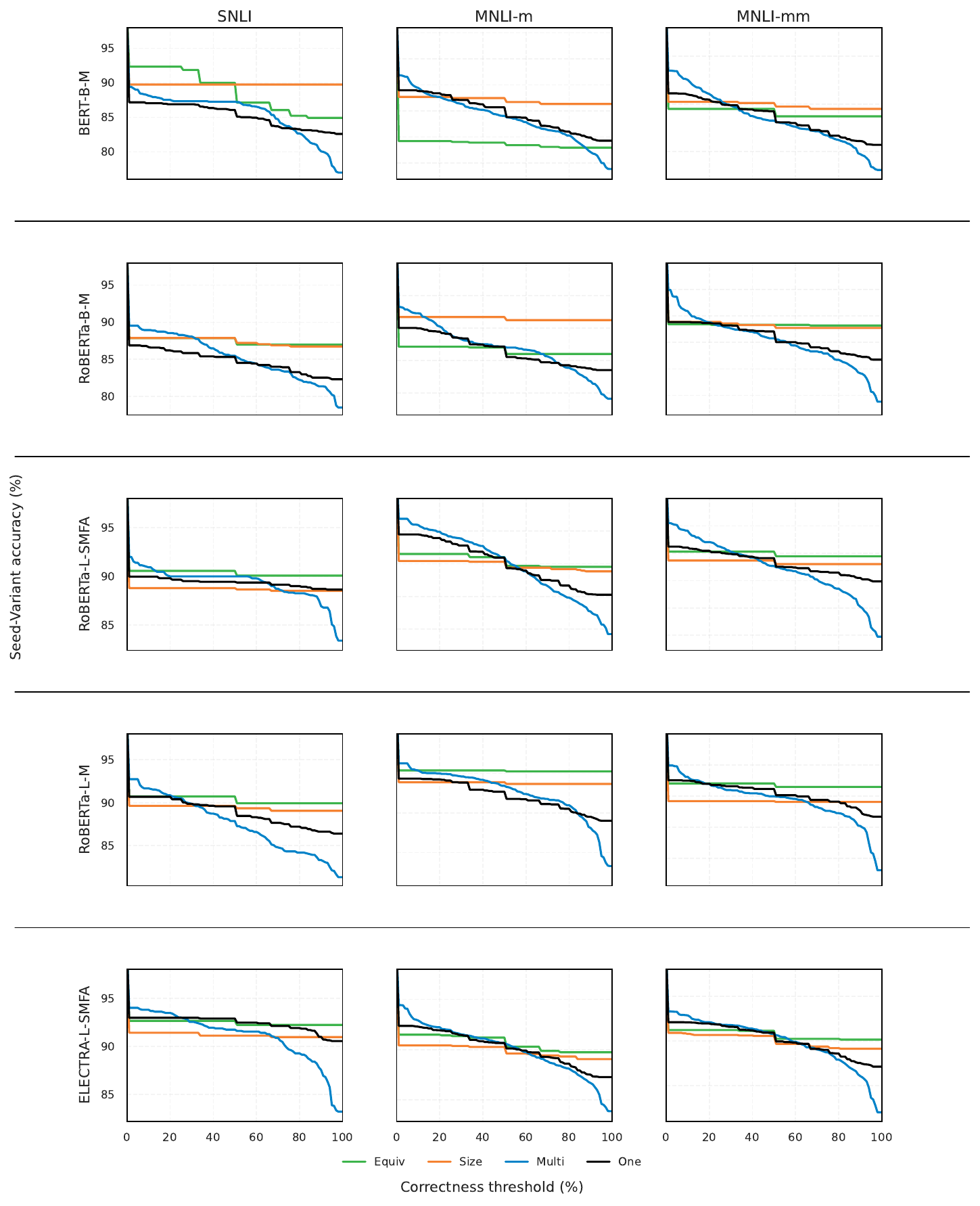}
    \caption{SV-acc curves across datasets for variants grouped by the origin model that suggested them (SNLI=150, MNLI-m=530, MNLI-mm=424). We only show the models fine-tuned on SMFA and MNLI, as SNLI fine-tuned models had lower scores and were less informative.}
    \label{fig:figure_origin_appendix}
\end{figure*}

\onecolumn
\begin{center}
\small
\begin{longtable}{llll}
\hline
\endfoot
\multicolumn{4}{c}{\textbf{SNLI}} \\
\midrule
\multicolumn{4}{p{\dimexpr\linewidth-2\tabcolsep\relax}}{\textbf{P:} A man pushing a hand-truck of boxes is bending over to pick up a pear.} \\*
\multicolumn{4}{p{\dimexpr\linewidth-2\tabcolsep\relax}}{\textbf{H:} A happy man is picking up a pear.} \\*
\textbf{Gold label:} E & \textbf{Correct label: \textcolor{red}{N}}  & \textbf{Annotations:} E(3) N(2) C(0) & \textbf{Avg rate:} 0.00 \\
\hline

\multicolumn{4}{p{\dimexpr\linewidth-2\tabcolsep\relax}}{\textbf{P:} A man in a colorful shirt and a lady in a white blouse sign copies of books for people.} \\*
\multicolumn{4}{p{\dimexpr\linewidth-2\tabcolsep\relax}}{\textbf{H:} Two people sign copies of their latest novel.} \\*
\textbf{Gold label:} E & \textbf{Correct label: \textcolor{red}{N}}  & \textbf{Annotations:} E(3) N(2) C(0) & \textbf{Avg rate:} 0.00 \\
\hline

\multicolumn{4}{p{\dimexpr\linewidth-2\tabcolsep\relax}}{\textbf{P:} A black dog is swimming with a ball in his mouth.} \\*
\multicolumn{4}{p{\dimexpr\linewidth-2\tabcolsep\relax}}{\textbf{H:} A black dog found a ball in the water and is bring it back to its owner.} \\*
\textbf{Gold label:} E & \textbf{Correct label:\textcolor{red}{N}}  & \textbf{Annotations:} E(3) N(2) C(0) & \textbf{Avg rate:} 0.00 \\
\hline

\multicolumn{4}{p{\dimexpr\linewidth-2\tabcolsep\relax}}{\textbf{P:} A wet child stands in chest deep ocean water.} \\*
\multicolumn{4}{p{\dimexpr\linewidth-2\tabcolsep\relax}}{\textbf{H:} The child s playing on the beach.} \\*
\textbf{Gold label:} E & \textbf{Correct label:\textcolor{red}{N}}  & \textbf{Annotations:} E(3) N(1) C(1) & \textbf{Avg rate:} 0.00 \\
\hline

\multicolumn{4}{p{\dimexpr\linewidth-2\tabcolsep\relax}}{\textbf{P:} The man in the brown shirt is holding the hand of the long-haired child in front of a painting.} \\*
\multicolumn{4}{p{\dimexpr\linewidth-2\tabcolsep\relax}}{\textbf{H:} A male has clothes on with his hand holding another young male in front of a painting.} \\*
\textbf{Gold label:} N & \textbf{Correct label:\textcolor{green}{N}}  & \textbf{Annotations:} E(2) N(3) C(0) & \textbf{Avg rate:} 0.00 \\
\hline

\multicolumn{4}{p{\dimexpr\linewidth-2\tabcolsep\relax}}{\textbf{P:} A mom and her boy are riding in a bumper car.} \\*
\multicolumn{4}{p{\dimexpr\linewidth-2\tabcolsep\relax}}{\textbf{H:} The mom and boy are at an amusement park.} \\*
\textbf{Gold label:} E & \textbf{Correct label:\textcolor{red}{N}}  & \textbf{Annotations:} E(3) N(2) C(0) & \textbf{Avg rate:} 0.00 \\
\hline

\multicolumn{4}{p{\dimexpr\linewidth-2\tabcolsep\relax}}{\textbf{P:} A man wearing a blue apron and long rubber boots is dragging a flotation device from a long row of flotation devices.} \\*
\multicolumn{4}{p{\dimexpr\linewidth-2\tabcolsep\relax}}{\textbf{H:} A man in rubber boots and a work apron is rubbing his face in between pulling floating objects.} \\*
\textbf{Gold label:} E & \textbf{Correct label:\textcolor{red}{N}}  & \textbf{Annotations:} E(3) N(1) C(1) & \textbf{Avg rate:} 0.00 \\
\hline

\multicolumn{4}{p{\dimexpr\linewidth-2\tabcolsep\relax}}{\textbf{P:} One girl sips a soda while another looks on, standing on a street in front of a bunch of bicycles.} \\*
\multicolumn{4}{p{\dimexpr\linewidth-2\tabcolsep\relax}}{\textbf{H:} A girl drinks a soda on the street in front of people} \\*
\textbf{Gold label:} N & \textbf{Correct label:\textcolor{green}{N}}  & \textbf{Annotations:} E(2) N(3) C(0) & \textbf{Avg rate:} 0.00 \\
\hline

\multicolumn{4}{p{\dimexpr\linewidth-2\tabcolsep\relax}}{\textbf{P:} people standing at a beach with Cameras.} \\*
\multicolumn{4}{p{\dimexpr\linewidth-2\tabcolsep\relax}}{\textbf{H:} A group of people standing at a beach filled with cameras.} \\*
\textbf{Gold label:} N & \textbf{Correct label:\textcolor{green}{N}}  & \textbf{Annotations:} E(2) N(3) C(0) & \textbf{Avg rate:} 0.00 \\
\hline

\multicolumn{4}{p{\dimexpr\linewidth-2\tabcolsep\relax}}{\textbf{P:} A woman holds a newspaper that says "Real change".} \\*
\multicolumn{4}{p{\dimexpr\linewidth-2\tabcolsep\relax}}{\textbf{H:} a woman on a street holding a newspaper that says "real change"} \\*
\textbf{Gold label:} E & \textbf{Correct label:\textcolor{red}{N}}  & \textbf{Annotations:} E(3) N(2) C(0) & \textbf{Avg rate:} 0.00 \\
\hline

\multicolumn{4}{p{\dimexpr\linewidth-2\tabcolsep\relax}}{\textbf{P:} A young man blew up balloons to craft into animals for the seven excited children that looked on.} \\*
\multicolumn{4}{p{\dimexpr\linewidth-2\tabcolsep\relax}}{\textbf{H:} The children watch the man make dogs and giraffes out of balloons} \\*
\textbf{Gold label:} E & \textbf{Correct label:\textcolor{red}{N}}  & \textbf{Annotations:} E(3) N(2) C(0) & \textbf{Avg rate:} 0.00 \\
\hline

\multicolumn{4}{p{\dimexpr\linewidth-2\tabcolsep\relax}}{\textbf{P:} A train conductor in coveralls is standing in the door of the trail.} \\
\multicolumn{4}{p{\dimexpr\linewidth-2\tabcolsep\relax}}{\textbf{H:} The conductor is walking in a field.} \\
\textbf{Gold label:} N & \textbf{Correct label:\textcolor{red}{C}}  & \textbf{Annotations:} E(0) N(3) C(2) & \textbf{Avg rate:} 0.00 \\
\hline

\multicolumn{4}{p{\dimexpr\linewidth-2\tabcolsep\relax}}{\textbf{P:} An older gentleman looks at the camera while he is building a deck.} \\
\multicolumn{4}{p{\dimexpr\linewidth-2\tabcolsep\relax}}{\textbf{H:} An older gentleman in overalls looks at the camera while he is building a stained red deck in front of a house.} \\
\textbf{Gold label:} E & \textbf{Correct label:\textcolor{red}{N}}  & \textbf{Annotations:} E(3) N(2) C(0) & \textbf{Avg rate:} 0.00 \\
\hline

\multicolumn{4}{p{\dimexpr\linewidth-2\tabcolsep\relax}}{\textbf{P:} Children, including one with a painted face, pet tiny turtles that are crawling in the green grass.} \\
\multicolumn{4}{p{\dimexpr\linewidth-2\tabcolsep\relax}}{\textbf{H:} Turtles are crawling in the white grass.} \\
\textbf{Gold label:} E & \textbf{Correct label:\textcolor{red}{N}}  & \textbf{Annotations:} E(3) N(0) C(2) & \textbf{Avg rate:} 0.00 \\
\hline

\multicolumn{4}{p{\dimexpr\linewidth-2\tabcolsep\relax}}{\textbf{P:} A group of people are in a rowboat in the ocean surrounded by seagulls.} \\
\multicolumn{4}{p{\dimexpr\linewidth-2\tabcolsep\relax}}{\textbf{H:} A bunch of people are in a wooden object on the water.} \\
\textbf{Gold label:} N & \textbf{Correct label:\textcolor{green}{N}}  & \textbf{Annotations:} E(2) N(3) C(0) & \textbf{Avg rate:} 0.00 \\
\hline

\multicolumn{4}{p{\dimexpr\linewidth-2\tabcolsep\relax}}{\textbf{P:} A man in a hard hat looks intimidated.} \\
\multicolumn{4}{p{\dimexpr\linewidth-2\tabcolsep\relax}}{\textbf{H:} He is working in a potentially dangerous field that requires a hard hat.} \\
\textbf{Gold label:} E & \textbf{Correct label:\textcolor{red}{N}}  & \textbf{Annotations:} E(4) N(1) C(0) & \textbf{Avg rate:} 0.00 \\
\hline
\midrule
\multicolumn{4}{c}{\textbf{MNLI-m}} \\
\midrule
 
\multicolumn{4}{p{\dimexpr\linewidth-2\tabcolsep\relax}}{\textbf{P:} Robust  came in third among words and phrases submitted (220 citations in the CR ), and unlike the previous two, it seems to be a genuinely new cliche; at any rate, Chatterbox hadn't previously been aware of its overuse.} \\
\multicolumn{4}{p{\dimexpr\linewidth-2\tabcolsep\relax}}{\textbf{H:} Robust is a legitimately new cliche unlike its predecessors.} \\
\textbf{Gold label:} N & \textbf{Correct label:\textcolor{red}{E}} & \textbf{Annotations:} E(2) N(3) C(0) & \textbf{Avg rate:} 0.00 \\
\hline

\multicolumn{4}{p{\dimexpr\linewidth-2\tabcolsep\relax}}{\textbf{P:} 3) The gap between the productivity of women and the productivity of men.} \\
\multicolumn{4}{p{\dimexpr\linewidth-2\tabcolsep\relax}}{\textbf{H:} The gap of genders.} \\
\textbf{Gold label:} N & \textbf{Correct label: \textcolor{green}{N}}  & \textbf{Annotations:} E(2) N(3) C(0) & \textbf{Avg rate:} 0.00 \\
\hline

\multicolumn{4}{p{\dimexpr\linewidth-2\tabcolsep\relax}}{\textbf{P:} A 1994 Roper Poll concluded that the NewsHour is perceived by the public as the most credible newscast in the country.} \\
\multicolumn{4}{p{\dimexpr\linewidth-2\tabcolsep\relax}}{\textbf{H:} A 1984 Poll concluded NewsHour is seen as the most credible newscast by the public.} \\
\textbf{Gold label:} E & \textbf{Correct label: \textcolor{red}{E}}  & \textbf{Annotations:} E(3) N(0) C(2) & \textbf{Avg rate:} 0.00 \\
\hline

\multicolumn{4}{p{\dimexpr\linewidth-2\tabcolsep\relax}}{\textbf{P:} Once the pious devotions are over, however, wine flows, fireworks explode, espetada (kebab) stalls flourish, and Monte regains normality for another 363 days.} \\
\multicolumn{4}{p{\dimexpr\linewidth-2\tabcolsep\relax}}{\textbf{H:} Monte is a normal location, outside of a period of pious devotion.} \\
\textbf{Gold label:} N & \textbf{Correct label:\textcolor{red}{E}}  & \textbf{Annotations:} E(2) N(3) C(0) & \textbf{Avg rate:} 0.00 \\
\hline
\multicolumn{4}{p{\dimexpr\linewidth-2\tabcolsep\relax}}{\textbf{P:} Christ on a crutch, what does he have to do to lose your support, stab David Geffen with a kitchen knife?} \\
\multicolumn{4}{p{\dimexpr\linewidth-2\tabcolsep\relax}}{\textbf{H:} Your support is unwavering.} \\
\textbf{Gold label:} E & \textbf{Correct label: \textcolor{green}{E}}   & \textbf{Annotations:} E(3) N(1) C(1) & \textbf{Avg rate:} 0.00 \\
\hline
\midrule
\multicolumn{4}{c}{\textbf{MNLI-mm}} \\
\midrule
\multicolumn{4}{p{\dimexpr\linewidth-2\tabcolsep\relax}}{\textbf{P:} Hani Hanjour, assigned to seat 1B (first class), soon followed.} \\
\multicolumn{4}{p{\dimexpr\linewidth-2\tabcolsep\relax}}{\textbf{H:} Bob was assigned to seat 2a} \\
\textbf{Gold label:} N & \textbf{Correct label:\textcolor{green}{E}}  & \textbf{Annotations:} E(0) N(4) C(1) & \textbf{Avg rate:} 0.00 \\
\hline

\multicolumn{4}{p{\dimexpr\linewidth-2\tabcolsep\relax}}{\textbf{P:} For example, even at age 2, they waited patiently to open a small gift until a guest had departed'proper etiquette in Chinese culture.} \\
\multicolumn{4}{p{\dimexpr\linewidth-2\tabcolsep\relax}}{\textbf{H:} They waited to open a small gift when they were very small because they'd been taught to do so.} \\
\textbf{Gold label:} E & \textbf{Correct label:\textcolor{green}{E}}  & \textbf{Annotations:} E(3) N(2) C(0) & \textbf{Avg rate:} 0.00 \\
\hline

\multicolumn{4}{p{\dimexpr\linewidth-2\tabcolsep\relax}}{\textbf{P:} . . . You got a conflict on that direction?} \\
\multicolumn{4}{p{\dimexpr\linewidth-2\tabcolsep\relax}}{\textbf{H:} No one brought up a varying opinion on that direction.} \\
\textbf{Gold label:} N & \textbf{Correct label:\textcolor{green}{N}}  & \textbf{Annotations:} E(0) N(3) C(2) & \textbf{Avg rate:} 0.00 \\
\hline

\multicolumn{4}{p{\dimexpr\linewidth-2\tabcolsep\relax}}{\textbf{P:} This information infrastructure proved most beneficial when it gave Penney's major vendors access to sales data via direct broadcast satellite.} \\
\multicolumn{4}{p{\dimexpr\linewidth-2\tabcolsep\relax}}{\textbf{H:} This infrastructure of information was used in order to provide customers with information regarding sales.} \\
\textbf{Gold label:} C & \textbf{Correct label:\textcolor{red}{E}}  & \textbf{Annotations:} E(0) N(1) C(4) & \textbf{Avg rate:} 0.00 \\
\hline

\multicolumn{4}{p{\dimexpr\linewidth-2\tabcolsep\relax}}{\textbf{P:} The operator must first select the work to be done, put aside the tickets that indicate she performed the sewing appropriate for those bundles and should be paid at the specified rate for the job, open the appropriate bundles, and position the pieces to be joined on the sewing table in preparation for sewing.} \\
\multicolumn{4}{p{\dimexpr\linewidth-2\tabcolsep\relax}}{\textbf{H:} Some operators prefer to set aside the tickets first.} \\
\textbf{Gold label:} C & \textbf{Correct label:\textcolor{red}{N}}  & \textbf{Annotations:} E(0) N(2) C(3) & \textbf{Avg rate:} 0.00 \\
\hline

\multicolumn{4}{p{\dimexpr\linewidth-2\tabcolsep\relax}}{\textbf{P:} However, we recognize that contributions at the Maennerchor Society level are not possible for all.} \\
\multicolumn{4}{p{\dimexpr\linewidth-2\tabcolsep\relax}}{\textbf{H:} Maennerchor Society level contributions are impossible for most everyone.} \\
\textbf{Gold label:} C & \textbf{Correct label:\textcolor{red}{N}}  & \textbf{Annotations:} E(0) N(1) C(4) & \textbf{Avg rate:} 0.00 \\
\hline

\multicolumn{4}{p{\dimexpr\linewidth-2\tabcolsep\relax}}{\textbf{P:} They're actually both teachers.} \\
\multicolumn{4}{p{\dimexpr\linewidth-2\tabcolsep\relax}}{\textbf{H:} The are teachers together.} \\
\textbf{Gold label:} N & \textbf{Correct label:\textcolor{green}{N}}  & \textbf{Annotations:} E(2) N(3) C(0) & \textbf{Avg rate:} 0.00 \\
\hline

\multicolumn{4}{p{\dimexpr\linewidth-2\tabcolsep\relax}}{\textbf{P:} Even though we receive operating funds from the state, there are a myriad of additional expenses to be met, such as welding equipment for sculpture, pottery wheels for ceramics, and computers for graphics.} \\
\multicolumn{4}{p{\dimexpr\linewidth-2\tabcolsep\relax}}{\textbf{H:} The state won't fund welding equipment, pottery wheels or computers.} \\
\textbf{Gold label:} E & \textbf{Correct label:\textcolor{red}{N}}  & \textbf{Annotations:} E(3) N(1) C(1) & \textbf{Avg rate:} 0.00 \\
\hline

\multicolumn{4}{p{\dimexpr\linewidth-2\tabcolsep\relax}}{\textbf{P:} Business units adopting both bar codes and EDI are therefore able to reduce the transaction costs for processing information about sales and orders.} \\
\multicolumn{4}{p{\dimexpr\linewidth-2\tabcolsep\relax}}{\textbf{H:} Business units use either bar codes or EDI.} \\
\textbf{Gold label:} N & \textbf{Correct label:\textcolor{green}{N}}  & \textbf{Annotations:} E(0) N(4) C(1) & \textbf{Avg rate:} 0.00 \\
\hline

\multicolumn{4}{p{\dimexpr\linewidth-2\tabcolsep\relax}}{\textbf{P:} Did your dad read to you too?} \\
\multicolumn{4}{p{\dimexpr\linewidth-2\tabcolsep\relax}}{\textbf{H:} Did someone read to you?} \\
\textbf{Gold label:} N & \textbf{Correct label:\textcolor{red}{E}}  & \textbf{Annotations:} E(1) N(4) C(0) & \textbf{Avg rate:} 0.00 \\
\hline

\multicolumn{4}{p{\dimexpr\linewidth-2\tabcolsep\relax}}{\textbf{P:} But he places most blame on how contemporary children are reared.} \\
\multicolumn{4}{p{\dimexpr\linewidth-2\tabcolsep\relax}}{\textbf{H:} People put the most blame on how contemporary children are reared.} \\
\textbf{Gold label:} N & \textbf{Correct label:\textcolor{green}{N}}  & \textbf{Annotations:} E(1) N(4) C(0) & \textbf{Avg rate:} 0.00 \\
\hline

\multicolumn{4}{p{\dimexpr\linewidth-2\tabcolsep\relax}}{\textbf{P:} I forgot about comic books!} \\
\multicolumn{4}{p{\dimexpr\linewidth-2\tabcolsep\relax}}{\textbf{H:} I did not mention comic books yet.} \\
\textbf{Gold label:} N & \textbf{Correct label:\textcolor{green}{N}}  & \textbf{Annotations:} E(1) N(3) C(1) & \textbf{Avg rate:} 0.00 \\
\hline

\caption{The original SNLI and MNLI-m/-mm problems whose variants were most poorly classified by all NLI models. We also considered LLMs, but their sub-selected seed did not include these problems. \textbf{Avg rate} represents the average accuracy of NLI models across all variants per seed/original problem. \textbf{Annotations} reports the annotation labels of 5 annotators from the SNLI and MNLI data, where the gold label is selected based on majority voting. \textbf{Correct label} is an inference we thought was correct. We are aware of the inherent disagreements in NLI labeling \cite{pavlick-kwiatkowski-2019-inherent}, but we refer to our corrected labels as the most likely labels. Note that we also further looked into the problems that were classified with $<\!5$\% average accuracy across all models (SNLI=27, MNLI-m=29, MNLI-mm= 24), out of which only 13  for SNLI, and MNLI-m, and 10 for MNLI-mm had a correct gold label.}
\label{tab:hardets_nli_problems} 
\end{longtable}
\end{center}
\twocolumn
\end{document}